\documentclass[11pt]{article}

\usepackage[preprint]{acl}
\usepackage{times}
\usepackage{latexsym}

\usepackage[T1]{fontenc}
\usepackage[utf8]{inputenc}

\usepackage{microtype}

\usepackage{inconsolata}

\usepackage{graphicx}

\usepackage{hyperref}
\usepackage{url}
\usepackage{booktabs}
\usepackage{tabularx}
\usepackage{multirow}
\usepackage{subcaption} % add this in preamble
\usepackage{wrapfig}
\usepackage{lineno}
\usepackage{algorithm}
\usepackage{algpseudocode}
\usepackage{amsmath}
\usepackage[table]{xcolor}
\usepackage{arydshln}
\usepackage{slnotation}
\usepackage{fdiaz}

\newcommand\freefootnote[1]{%
  \let\svthefootnote\thefootnote% Save the current footnote number
  \let\thefootnote\relax% Make \thefootnote do nothing so \footnotetext doesn't increment the footnote counter
  \footnotetext{\textsuperscript{*}#1}% Add the footnote text with an asterisk
  \let\thefootnote\svthefootnote% Restore the original footnote behavior
}

\usepackage[most]{tcolorbox}
\usepackage{listings}
\tcbuselibrary{breakable}

\newtcolorbox{promptbox}{
    colback=gray!5, 
    colframe=black!75, 
    fontupper=\small\ttfamily, 
    sharp corners, 
    boxrule=0.5pt,
    breakable,       % Allows splitting across columns/pages
    halign=left,     % Ensures text hugs the left margin
    before skip=10pt,
    after skip=10pt
}

\newtcolorbox{systemprompt}{
    breakable, enhanced,
    colback=blue!4, colframe=blue!35,
    fontupper=\small\ttfamily,
    title={\footnotesize\sffamily\bfseries SYSTEM},
    attach boxed title to top left={yshift=-2mm, xshift=4mm},
    boxed title style={colback=blue!35, colframe=blue!35, sharp corners},
    sharp corners, boxrule=0.5pt,
    before skip=6pt, after skip=2pt
}

\newtcolorbox{userprompt}{
    breakable, enhanced,
    colback=teal!4, colframe=teal!40,
    fontupper=\small\ttfamily,
    title={\footnotesize\sffamily\bfseries USER},
    attach boxed title to top left={yshift=-2mm, xshift=4mm},
    boxed title style={colback=teal!40, colframe=teal!40, sharp corners},
    sharp corners, boxrule=0.5pt,
    before skip=2pt, after skip=10pt
}

\title{Multi-Agent Transactive Memory}

\author{
To Eun Kim\textsuperscript{1*} \quad 
Xuhong He\textsuperscript{1*} \quad 
Dishank Jain\textsuperscript{1*} \\
\textbf{Ambuj Agrawal\textsuperscript{1}} \quad 
\textbf{Negar Arabzadeh\textsuperscript{2}} \quad 
\textbf{Fernando Diaz\textsuperscript{1}} \\
\textsuperscript{1}Carnegie Mellon University \quad \textsuperscript{2}University of California, Berkeley
}

\begin{document}
\maketitle

% Notations used in the main file.
% Extension of slnotation.sty by the followings:
%   1. command name change 
%       by \let\newcommand\oldcommand
%       or \newcommand (when the original command has arguments)
%   2. value redefining 
%       by \renewcommand{\commandname}[number_of_arguments]{new definition}
%   3. introducing completely new commands
%       by \newcommand

% <General Rules>
%   representation space: mathcal{x}
%   instance in that space: lower case
%   set in that space: \textrm{upper case}
%   parameterized functions: lower case w parameters as subscript
%   parameters: greek letters

%%%%%%%%%
% Terms
%%%%%%%%%

%%%%%%%%%
% Basics
%%%%%%%%%
\renewcommand{\slReals}{\mathbb{R}}

%%%%%%%%%
% RAG
%%%%%%%%%

% ML basics - input ouput
\let\instance\mlInput  % x
\let\instanceSpace\mlInputs  % \mathcal{X}
\let\outputSpace\mlOutputs
\renewcommand{\mlOutput}{\hat{y}}
\let\output\mlOutput
\renewcommand{\mlTarget}{y}
\let\target\mlTarget

% RAG pipeline
\newcommand{\query}{q}
\newcommand{\querySpace}{\mathcal{Q}}

\newcommand{\rModel}{\mathcal{R}}

\newcommand{\llmModel}{\mathcal{G}}
\newcommand{\queryGenerator}{\phi_q}
\newcommand{\promptGenerator}{\phi_p}
\newcommand{\augmentedPrompt}{\overline{\instance}}

% Interim results (documents, rankings, retrieval scores)
\newcommand{\doc}{d}
\newcommand{\corpus}{\mathcal{D}}
\newcommand{\ltrtDataset}{\mathcal{S}}

\newcommand{\retrievalResult}{z}
\newcommand{\retrievalResultSpace}{\mathcal{Z}}

\let\ranking\raRanking

\newcommand{\retrievalScores}{\mathbf{s}}
\newcommand{\retrievalScore}{s}

% Evaluation
\newcommand{\metricThreshold}{\theta}
\newcommand{\utilityMetric}{\evalMetric_u}
\newcommand{\relevanceMetric}{\evalMetric_r}

%%%%%%%%%
% Agent, Environemtn
%%%%%%%%%
\newcommand{\agent}{A}
\newcommand{\agenti}{\agent_i}
\newcommand{\agentj}{\agent_j}
\newcommand{\agents}{\mathcal{\agent}}
\newcommand{\numagents}{n}

\newcommand{\env}{E}
\newcommand{\envi}{\env_i}
\newcommand{\envj}{\env_j}
\newcommand{\envs}{\mathcal{E}}
\newcommand{\numenvs}{m}

%%%%%%%%%
% Trajectories
%%%%%%%%%
\newcommand{\trajectory}{\mathcal{T}}
\newcommand{\reasoning}{\mathbf{r}}
\newcommand{\action}{\mathbf{a}}
\newcommand{\observation}{\mathbf{o}}

\newcommand{\sar}{\tau}
\newcommand{\sari}{\sar_i}
\newcommand{\sarj}{\sar_j}
\newcommand{\sart}{\sar_t}

\newcommand{\goal}{g}

\newcommand{\horizon}{H}

\newcommand{\winsize}{l}

%%%%%%%%%
% Rerankers
%%%%%%%%%
\newcommand{\svmrank}{SVMRank\xspace}
\newcommand{\ffn}{FFN\xspace}
\newcommand{\lambdamart}{LambdaMART\xspace}

%%%%%%%%%
% Commenting
%%%%%%%%%
\newcommand{\missingcitation}[0]{{\color{red}[CITE]}}
\newcommand{\tekcomment}[1]{\textcolor{red}{[TEK: #1]}}
\newcommand{\aacomment}[1]{\textcolor{blue}{[AA: #1]}}
\newcommand{\fdcomment}[1]{\textcolor{cyan}{[FD: #1]}}
\newcommand{\nacomment}[1]{\textcolor{magenta}{[NA: #1]}}

\begin{abstract}
Decentralized deployment of LLM agents with diverse capabilities motivates infrastructure for knowledge sharing across heterogeneous agent populations. Just as search engines index human-generated artifacts for human problem solving, retrieval systems can organize agent-generated artifacts for reuse across agents, extending retrieval-augmented generation from human-authored documents to agent-generated ones. Agent trajectories in particular encode reusable procedural knowledge, yet are typically discarded after a single use or retained only by the producing agent, forcing newly instantiated agents to rediscover existing solutions. We propose Multi-Agent Transactive Memory (MATM), a framework for population-level storage and retrieval of agent trajectories, where \textit{producer agents} contribute trajectories to a shared repository and \textit{consumer agents} retrieve them to improve task execution. In interactive environments with long, procedurally rich trajectories (ALFWorld and WebArena), retrieving from MATM improves downstream task performance and reduces interaction steps without coordination or joint training, positioning MATM as a design pattern for population-level experience sharing in open agent ecosystems.
\end{abstract}

% <Original>
% The decentralized deployment of LLM agents with diverse capabilities across diverse tasks motivates infrastructure for knowledge sharing across heterogeneous agent populations. Just as search engines index human-generated artifacts to support human problem solving, retrieval systems can organize agent-generated artifacts for reuse across agent populations. We extend retrieval-augmented generation---which demonstrates the value of human-authored artifacts to individual agents---to retrieval of agent-generated artifacts supporting a population of agents. In particular, agent trajectories encode reusable procedural knowledge, yet these artifacts are typically discarded after a single use or retained only by the producing agent, forcing newly instantiated agents to repeatedly rediscover existing solutions. We propose Multi-Agent Transactive Memory (MATM), a framework for population-level storage and retrieval of agent-generated trajectories, where \textit{producer agents} contribute trajectories to a shared repository and \textit{consumer agents} retrieve them to improve task execution. We focus on interactive environments (ALFWorld and WebArena), where trajectories are long and encode especially rich procedural structure. Our experiments demonstrate that retrieving trajectories from MATM improves downstream task performance and reduces interaction steps without coordination or joint training. These results position MATM as a design pattern for population-level experience sharing in open agent ecosystems.
\freefootnote{Denotes equal contribution.
% \\Correspondence: toeunk@cs.cmu.edu
}
\freefootnote{\url{https://github.com/kimdanny/matm}}
% \freefootnote{\url{https://huggingface.co/datasets/toeunkim/matm-trajectories}}
\begin{figure}[t]
  \includegraphics[width=\columnwidth]{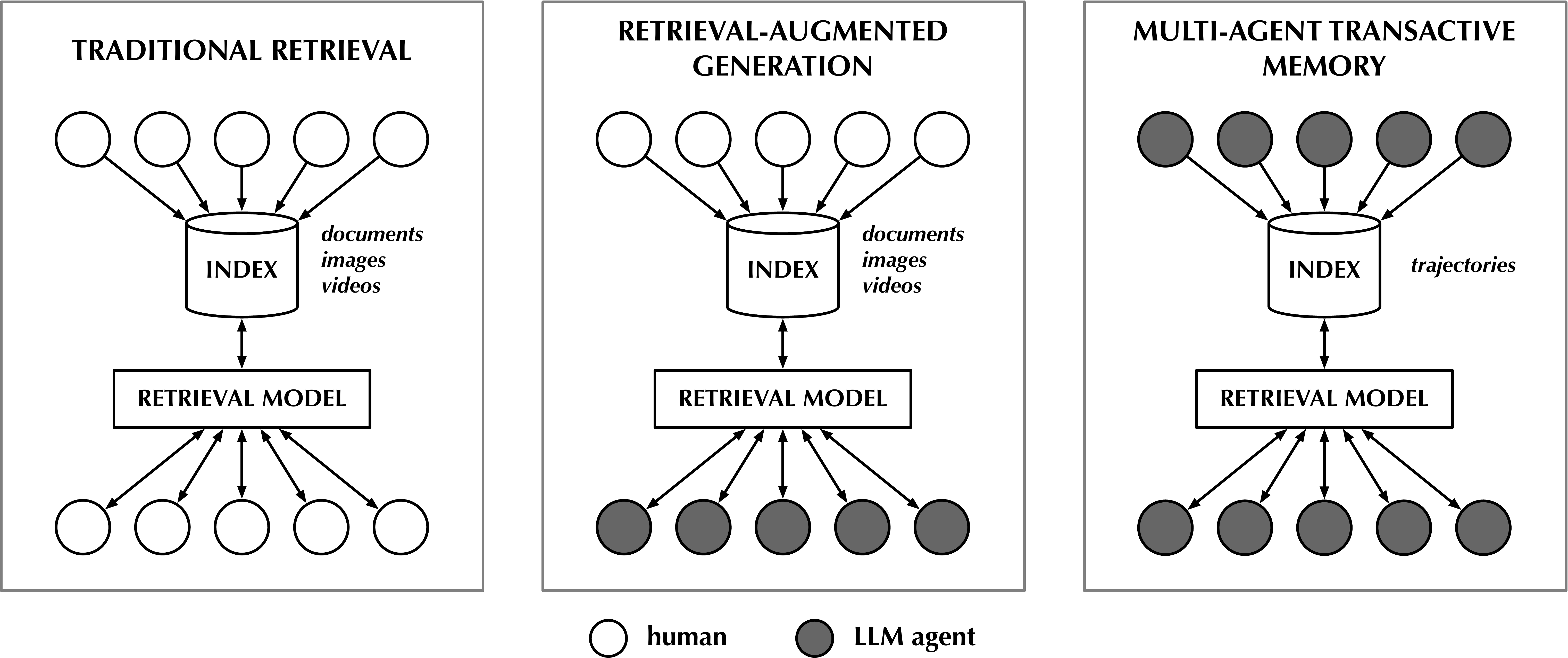}
  \caption{
  Multi-Agent Transactive Memory (MATM). Traditional search serves humans retrieving human-authored documents. RAG extends this to agents retrieving from human-generated corpora. MATM goes further, letting agents retrieve agent-generated artifacts such as interaction trajectories, which are atypical documents unlike human-written text. 
  }
  \label{fig:teaser}
\end{figure}

\section{Introduction}\label{sec:introduction}

% 1. WHAT IS THE PROBLEM
As heterogeneous LLM agents are deployed across increasingly diverse domains, research on individual agent must be complemented by methods for supporting decentralized populations of agents.  The need for population-level infrastructure \citep{jordan2025collectivist, hammond2025multi, tomavsev2025distributional} has motivated protocols to support agent-tool interaction \citep{mcp2026} as well as inter-agent communication \citep{a2a_protocol2026}. Beyond standards, tools such as search engines are beginning to be optimized for agents \citep{zamani:reml}.  
Just as search engines and retrieval-augmented generation (RAG) index human-generated artifacts for human and agent problem solving, retrieval systems can organize agent-generated artifacts for reuse across agent populations (Figure \ref{fig:teaser}).

% 2. WHY IS IT IMPORTANT
Artifact sharing and reuse are essential for scalable, efficient, and continually improving agent populations. As agents operate across environments, they produce intermediate artifacts that contain rich procedural knowledge, such as action-observation trajectories \citep{muennighoff-etal-2025-s1}, and the ability to reuse learned behaviors and continually acquire new experience is critical for scalability and long-term performance \citep{wang2025agentworkflow, liang2026skillnet, shi2025continual}. In contrast with RAG, agent-generated artifacts can be more suitable for agent consumption than human-authored documents \citep{chen2026agentir}. Reuse is further motivated by cost: many modern agents rely on inference-time scaling, so reusing their intermediate artifacts reduces redundant reasoning and exploration \citep{kaplan2020scaling,yao2023tree,wu2024scaling,welleck2024from,ahmed2025retrieval}.

% 3. WHY ARE CURRENT APPROACHES INSUFFICIENT
Existing approaches to artifact reuse are insufficient for heterogeneous agent ecosystems. Prior work on reasoning or thought reuse \citep{zheng2024synapse, ouyang2025reasoningbank, ahmed2025retrieval} improves cost-efficiency and effectiveness within individual agents, but reuse remains limited to the original artifact-producer; despite substantial overlap in the tasks agents solve, interaction trajectories are typically discarded after a single use \citep{zheng2024synapse, zhao2024expel}, causing newly instantiated agents to rediscover solutions that already exist elsewhere.
Related paradigms such as transfer learning \citep{konidaris:portable-options, brunskill:multi-task-rl-sample-complexity} and knowledge distillation \citep{li2025naturalthoughts, kang2025distilling} require alignment between source and target domains and often demand additional training, making them impractical for dynamically instantiated heterogeneous populations. Centralized multi-agent coordination methods \citep{dang2025multiagent} further assume \textit{cooperative settings} and shared protocols, constraining their applicability in open ecosystems \citep{tranMultiAgentCollaborationMechanisms2025} where agents freely join.  Indeed, based on analysis of Moltbook, \citet{liDoesSocializationEmerge2026} identify shared social memory as a missing prerequisite for the development of agent societies.
More broadly, the population is increasingly seen as the necessary unit of analysis: intelligence is fundamentally collective \citep{jordan2025collectivist}, heterogeneous agent populations are already being deployed with information asymmetry as a central challenge \citep{hammond2025multi}, and shared experience repositories require mechanisms that separate high-utility contributions from redundant ones \citep{tomavsev2025distributional}---a substrate that remains missing.

% 4. WHAT DO WE PROPOSE
To address this gap, we propose Multi-Agent Transactive Memory (MATM), a framework for population-level storage and retrieval of agent-generated artifacts, based on the concept of transactive memory \citep{wegner1987transactive}, in which human groups coordinate by distributing knowledge across individuals by using shared mechanisms for locating and retrieving relevant information. Similarly, MATM maintains a shared repository to which agents can contribute artifacts produced during their own task execution (\textit{producer agents}) and from which agents can retrieve procedural knowledge to improve their own task effectiveness and efficiency (\textit{consumer agents}). Roles are not mutually exclusive: an agent may produce trajectories in one context and consume them in another.

This producer-consumer structure induces a two-sided marketplace for agent-generated procedural knowledge, with clear attribution between retrieved artifacts and their sources. As more agents interact with the repository, MATM grows organically, accumulating a corpus across an increasingly diverse set of tasks and environments. Operating as a specialized retrieval system over agent-generated artifacts, MATM further enables retrieval functions that go beyond generic similarity search, including agent-specific personalization, producer trust modeling, and periodic update of retriever as the population evolves.

% 5. WHAT DO WE SHOW
We empirically demonstrate the effectiveness of MATM in interactive environments (ALFWorld \citep{shridhar2021alfworld} and WebArena \citep{zhou2024webarena}). 
We first show that agents consistently benefit from a simple single-stage retrieval pipeline: retrieving relevant trajectories from a MATM repository populated by diverse agents not only improves downstream task performance without requiring additional coordination or joint training, but also improves task efficiency as measured by a reduced number of interaction steps. 
We further introduce an efficient yet powerful learning-to-rank (LTR)-based trajectory reranking stage. With simple featurization of trajectory information, reranking yields better retrieval quality, leading to improved task effectiveness and greater step efficiency. 
Moreover, we find that retrieval benefit extends to both weaker and stronger agents, generalizes across tasks, and continues to improve as the repository grows.
Taken together, our contribution is focused on the population-level formulation of transactive memory and the findings it makes measurable: built on deliberately simple yet strong retrieval components, MATM provides a scalable mechanism for population-level experience reuse that lets agents leverage collective trajectories rather than repeatedly rediscovering solutions in isolation.

\section{Related Work (Appendix \ref{sec:app:related-work})}
Memory has long played a role in the development of AI agents. Existing approaches can be understood as memory over various sources of data. 

\textit{Memory of training data} provides agents with access to knowledge explicitly or implicitly stored during optimization.   Explicit methods include nearest-neighbor algorithms \citep{cover-hart:nn,khandelwal:knnlm} or case-based reasoning \citep{kolodner:intro-cbr,das:nl-cbr} or implicit behaviors \citep{carlini:memorization}.  

\textit{Memory over experience data} provides agents with access to traces of their own interactions.  Historically, methods reflecting memory of experience data include early cognitive architectures like SOAR \citep{laird:soar}, reinforcement learning \citep{lin:experience-replay},  and neural networks \citep{weston:memory-networks}. In the context of LLM agents, recent extensions treat an agent's own interaction history as retrievable context, giving rise to memory-augmented generation where past conversations or execution traces guide future behavior \citep{shinn2023reflexion, majumder2024clin, zheng2024synapse}. 

Agents generate rich intermediate artifacts during problem solving---action-observation trajectories, thinking traces, plans, workflows, and reusable code analogous to options in reinforcement learning \citep{Garcia19compressionMacro, veeriah2021discovery}. At the trajectory level, prior work retrieves reasoning templates \citep{yang2024buffer, ahmed2025retrieval} or reuses action-observation trajectories for downstream decisions \citep{zheng2024synapse, zhao2024expel}, while other work extracts and reuses more abstract artifacts such as causal abstractions, workflows, skills, and code \citep{majumder2024clin, wang2024voyager, wang2025agentworkflow, klissarov2025maestromotif, wang2025inducing, ouyang2025reasoningbank, arabzadeh2026thinkingtrace}.
Other systems further organize such memory through agentic \citep{xu2025amem}, hierarchical multi-agent team \citep{zhang2025gmemory}, or self-evolving \citep{wu2026evolveR} structures, yet all remain confined to a single agent's or a single team's own experience.
Emerging shared repositories of agent-generated artifacts \citep{schmidgall2025agentrxiv} move beyond this by pooling outputs across agents, but have so far been studied among agents pursuing a common objective rather than heterogeneous agents across diverse tasks.
Artifacts can also serve as distillation signals across models \citep{yang2025supercorrect, li2025naturalthoughts, kang2025distilling}, suggesting that a population-level repository such as MATM can also supply training signals for distillation.

\textit{Memory of external data} provides agent with access to shared artifact repositories and is represented by retrieval-augmented generation (RAG) \citep{lewis2020retrieval}, which enhances language models by conditioning generation on retrieved external context \citep{fan2024survey}.

\section{Multi-Agent Transactive Memory}
We propose a population-level memory of experience data. Rather than private per-agent memory, MATM is a shared, structured repository that heterogeneous agents both contribute to and retrieve from. This shifts artifact reuse from an individual optimization mechanism to a collective knowledge infrastructure, enabling continual learning and cross-agent transfer, reducing redundant exploration, and supporting cumulative capability growth at the population level.

% \paragraph{Task Environment.}
Formally, we consider a population of $\numagents$ LLM agents $\agents = \{ \agenti\}_{i=1}^\numagents$ each potentially pursuing heterogeneous goals and operating across one or more environments $\envs = \{ \envi \}_{i=1}^\numenvs$.
A task is specified by a description $\instance \in \instanceSpace$, which corresponds to goal specification or initial state for an agent.
Given a task description $\instance$ in environment $\envi$, an LLM agent $\agentj$ performs a series of interleaved turns with the environment to solve the task. During this process, we can record a variable-length trajectory $\trajectory_{\envi, \agentj} = (\sart)_{t=1}^{\horizon}$, where each step $\sart$ represents a unit of interaction. For example, in a web navigation environment, each $\sart$ corresponds to an action-observation pair (e.g., a click action and the resulting HTML observation) in the interaction sequence. For simplicity, we denote this agent-generated trajectory as $\trajectory$.

As the agent population $\agents$ operates across environments $\envs$, these trajectories accumulate into a rich collection of intermediate artifacts. We denote the population-level artifact repository as $\corpus = \{ \trajectory \}$ and refer to this incrementally growing shared memory as Multi-Agent Transactive Memory (MATM). 
Within this framework, we refer to agents that contribute trajectories to $\corpus$ as \textit{producer agents} and those that retrieve from it to aid their own task-solving as \textit{consumer agents}, where these roles are not mutually exclusive.
Our goal is to study how retrieval from this population-level memory can be optimized to improve outcomes for the population of consumer agents.

%>>>>>>Original
Although we focus on raw trajectories, this does not preclude higher-level abstractions such as skills or induced policies.
We use trajectories as the lowest-level, most universally available outputs, which still allow higher-level abstractions \citep{klissarov2025maestromotif} to be built on top. 
Moreover, retrieval over interactive trajectories is itself non-trivial. Unlike prompt-based artifacts such as \texttt{SKILL.md} files, which can be indexed with standard RAG techniques \citep{liang2026skillnet} or transformed into more retrieval-friendly forms \citep{arabzadeh2026thinkingtrace}, state-conditioned retrieval over action-observation histories has received much less attention and is the setting we study.

%%%%%%%%%%%%%%%%%

\subsection{Transactive Memory Indexing \& Retrieval}\label{subsec:matm-indexing-retrieval}
For action-observation trajectories, we adopt a state-conditioned key-value indexing scheme in which recent interaction history serves as the retrieval key and the subsequent interaction segment as the stored value. This allows consumer agents to retrieve continued guidance conditioned on their current state rather than only on the original task instruction.

Let each step $\sar_t = (\observation_t, \action_t)$ pair the observation $\observation_t$ seen by the agent with the action $\action_t$ taken in response.
Given a window size $\winsize$, we construct a key-value pair for each interaction step $t$.
The key is an embedding of the task description, the preceding $\winsize$ steps, and the current observation:
\[
\mathbf{e}_{\text{key}}^{(t)} = f\left(\instance, \sar_{t-\winsize}, \ldots, \sar_{t-1}, \observation_t\right),
\]
where $f$ is a shared embedding function.
The corresponding value consists of the $\winsize$ steps starting at step $t$:
\[
\left(\sar_t, \ldots, \sar_{t+\winsize-1}\right).
\]
This value serves as the document $\doc$ retrieved by an agent at inference time.
Both the preceding and subsequent windows are truncated at trajectory boundaries.
% Let each step $\sart = (\observation_t, \action_t)$ pair the observation $\observation_t$ the agent sees with the action $\action_t$ it takes in response.
% Given a window size $\winsize$, for each interaction step $t$ we define the key as the task description, the previous $\winsize$ steps, and the current observation,
% $\mathbf{e}_{\text{key}}^{(t)}=f\left(\instance,\sar_{t-\winsize},\ldots,\sar_{t-1},\observation_t\right)$,
% and the associated value as the $\winsize$ steps starting from the current one,
% $\left(\sar_{t},\ldots,\sar_{t+\winsize-1}\right)$,
% which serves as the document $\doc$ that an agent retrieves at inference time, where $f$ is a shared embedding function. Both windows are truncated at trajectory boundaries.
The key thus mirrors exactly what a consumer agent observes before acting at step $t$, while the value begins with the action $\action_t$ taken in that state, so the actions it provides are the $\winsize$ actions that follow the key.
Because this stored value serves as procedural guidance for the consumer's next steps, we index successful continuations; reusing failed trajectories as negative signal is a complementary mechanism we leave to future work.

Given a task description $\instance$ and MATM memory $\corpus$, a trajectory retriever $\rModel$ forms a search query $\query$, following the process described above and returns a ranked list $\ranking = \rModel(\instance, \corpus, K)$ of candidate trajectory chunks, where higher-ranked chunks are predicted to be more relevant for the current task and state.
The trajectory retriever $\rModel$ may be instantiated as a dense retriever, or a cascaded retrieval pipeline combining an initial retriever with a reranker.

Although the embedding model $f$ can in principle be tuned to better support artifact retrieval, we instead explore a simpler and underexamined approach that aligns retrieval results with consumer-agent preferences using lightweight learning-to-rank (LTR) rerankers \citep{cao2007ltr}. 

%%%%%%% LTRT
\subsection{Learning To Rank Trajectories (LTRT)}\label{subsec:ltrt}
Learning to rank pipelines consist of a retrieval stage known as candidate generation, followed by a feature-based ranking stage that re-orders the retrieved candidates.  

A feature map $\phi$ is designed to capture multiple complementary aspects of trajectory usefulness. 
Let $\phi(\query, \doc) \in \slReals^{z}$ be a feature map that extracts features by inspecting the query $\query$ and document key $\doc$. Let $g_\theta: \slReals^{z} \mapsto \slReals$ be a parameterized scoring function, where larger outputs indicate greater predicted helpfulness of a document $\doc$ value for task $\query$.

In MATM, we define features in six categories (Appendix~\ref{sec:app:ltr-features}):
\begin{inlinelist}
    \item producer agent metadata (e.g.,relevant benchmark scores);
    \item consumer agent metadata (e.g., agent ID);
    \item first-stage retrieval features (e.g., retrieval scores);
    \item query features (e.g., query length);
    \item trajectory features (e.g., trajectory length); and 
    \item query-trajectory interaction features (e.g., query-trajectory embedding similarity).
\end{inlinelist}
Two of these categories carry particular conceptual weight. Producer-agent metadata is designed to enable a form of \textit{trust modeling}, allowing the reranker to learn which producers' trajectories are reliable for a given context. Consumer-agent metadata is designed to enable \textit{personalization} of retrieval to the individual consumer that has joined the MATM framework, since the same trajectory may be more or less useful depending on the consumer's capabilities.

Training $g_\theta$ requires supervision over which retrieved trajectories actually help. Rather than treating relevance as semantic similarity, we label trajectory chunks by their \textit{marginal utility} \citep{salemiEvaluatingRetrievalQuality2024}: a chunk is helpful to the extent that injecting it into the consumer agent improves task outcome relative to running the same agent with no retrieval. The concrete procedure for collecting these labels is intertwined with how the memory itself is built, and we describe both jointly in Section~\ref{subsec:matm-construction} and \ref{subsec:ltrt-dataset}.

\section{Experimental Setup}\label{sec:exp-setup}

We instantiate MATM in two interactive benchmarks: ALFWorld \citep{shridhar2021alfworld}, a text-based household-task environment, and WebArena \citep{zhou2024webarena}, a web navigation-based task environment.
Each benchmark yields its own MATM index, populated by trajectories from 35 to 37 producer agents and consumed by 34 consumer agents (full population list in Appendix \ref{sec:app:population-list}).

For ALFWorld, we use the official train and test split, using 3553 episodes from the official training set, and evaluating on all 274 official test episodes (Appendix \ref{sec:app:alfworld-split}). For WebArena, which ships without a standard train/test partition, we construct a custom split that preserves the distribution of task intents, yielding 724 training and 88 test episodes (Appendix \ref{sec:app:webarena-split}). In both benchmarks, all MATM construction and LTRT training is performed strictly on the training partition, so the test set remains untouched throughout the MATM corpus construction phase. As a result, the test set may contain questions whose task type or environment configuration overlaps with those seen during MATM construction (e.g., similar map layouts in ALFWorld or shared website domains in WebArena), but no test question is itself solved by a producer agent and inserted into the corpus.

%%%%%%%%%%%%%%%%%
\subsection{Transactive Memory Construction}\label{subsec:matm-construction}
To construct MATM emerging as a trajectory storage of a population of agents producing and consuming trajectories, we expand the MATM corpus through two phases. \textit{Pre-population} initializes the index from existing trajectory sources, and \textit{incremental update} grows it as the producer and consumer agent population processes new training questions and contributes \textit{successful} trajectories back to the shared memory. Both phases operate exclusively over the training partition.

\paragraph{Pre-Population.}
The pre-population phase seeds an initial index $\corpus_0$ with publicly available trajectories. For ALFWorld, we collect trajectories from a trained seq2seq model released by the benchmark authors, supplemented by trajectories generated by running Qwen3-32B and GPT-OSS 20B on the training set. For WebArena, we collect publicly available trajectories produced by GPT-4-Turbo, GPT-4-Turbo-Preview, and Claude-3.5-Sonnet from the official benchmark runs, again supplemented by Qwen3-32B and GPT-OSS 20B trajectories generated on the training set.
In both cases, the collected trajectories are segmented into document chunks, encoded with the shared embedding function $f$, and inserted into the dense index $\corpus_0$, yielding 85,615 and 8,547 chunks for ALFWorld and WebArena respectively.

\paragraph{Incremental Update.}
After pre-population, MATM grows incrementally as the agent population operates over a stream of new training questions. This phase serves a dual purpose. It enriches the index with trajectories from a diverse set of producer agents, and it simultaneously creates the supervision signals needed to train LTRT rerankers.

The training questions are organized into partitions $\{\instanceSpace_p\}_{p=1}^{P}$ processed sequentially. Within each partition, every question is assigned to a producer agent via a deterministic allocation function $\sigma(\instance, \mathcal{\agent}, p)$ that ensures balanced coverage across task categories and agents (Appendix \ref{sec:app:allocation_function}). For each assigned pair $(\instance, \agent_n)$, the agent first attempts $\instance$ without retrieval to obtain a baseline trajectory $\trajectory_{\mathrm{base}}$ and score $s_{\mathrm{base}}$, which serves as the reference point for downstream marginal-utility comparisons. We then sample $T$ branching points $\{t_1, \dots, t_T\}$ randomly from the steps of $\trajectory_{\mathrm{base}}$. 
Following \citet{chang2015learning}, at each branching point $t$, we \textit{roll in} to the corresponding prefix $h_t = (\tau_1, \dots, \tau_t)$ and retrieve the top-$K$ chunks from the current index most similar to $\instance$ combined with $h_t$. We then \textit{roll out} $|\mathcal{I}|$ one-shot trajectory-augmented generations from $h_t$, one per selected rank $j \in \mathcal{I} \subseteq \{1,\ldots,K\}$, scoring each resulting trajectory $\trajectory_t^{(j)}$ as $s_t^{(j)}$.

The loop produces two outputs simultaneously. Any trajectory meeting a quality threshold $\metricThreshold$, including the baseline, is added to a trajectory buffer $\mathcal{B}_p$.
Here, $\metricThreshold$ thresholds the downstream task metric at collection time; for the binary-success benchmarks studied here (ALFWorld and WebArena) we set $\metricThreshold=1$.
After all questions in $\instanceSpace_p$ are processed, every trajectory in $\mathcal{B}_p$ is segmented, embedded with $f$, and added to the index, yielding $\corpus_p$. 

After all partitions are processed, the final MATM corpus contains 86,833 chunks for ALFWorld and 20,102 chunks for WebArena (Appendix~\ref{sec:app:index-stat}).\footnote{All trajectories are available at: \url{https://huggingface.co/datasets/toeunkim/matm-trajectories}.}
The full algorithm is given in Appendix~\ref{app:sec:data_construction}.

\subsection{LTRT Dataset \& Reranker Training}\label{subsec:ltrt-dataset}
The incremental construction procedure yields a labeled training dataset $\ltrtDataset = \{ (\query, \doc, \ell) \}$ for the LTRT reranker. 
For each retrieved chunk $\doc_t^{(j)}$ evaluated at branching point $t$, we record the tuple 
$(\query_t, \doc_t^{(j)}, \ell)$ with label $\ell = s_t^{(j)} - s_{\mathrm{base}}$, 
capturing the chunk's marginal utility relative to the no-retrieval baseline. 
With $Q = \sum_p |\instanceSpace_p|$ training questions, $T$ branching points per question, and $|\mathcal{I}|$ 
ranks evaluated per branching point, the resulting dataset contains 
$Q \times T \times |\mathcal{I}|$ labeled tuples.
In our experiments, we used $T=2$ for ALFWorld, $T=1$ for WebArena. We sample rank positions $\mathcal{I}=\{1, 5, 10, 15, 20\}$ for both benchmarks, exposing the LTRT model to candidates across the full retrieval depth while avoiding the cost of generating all twenty labeled episodes.
This yields $355\times2\times5=3{,}550$ labeled tuples for ALFWorld and $58\times1\times5=290$ for WebArena.

% \textbf{Features.} 
We compute 44 features per $(\query, \doc)$ pair, spanning the 
six categories introduced in \S\ref{subsec:ltrt}. The full feature list is provided in Appendix \ref{sec:app:ltr-features}.
% 
% \textbf{Reranker variants.} 
We train three reranker families spanning common LTR 
paradigms: a pointwise feed-forward network (FFN), pairwise 
LambdaMART~\citep{wu2010adapting}, and pairwise SVMRank~\citep{joachims2006training}. 20\% of the training set was used for the validation set for LTRT training.

%%%%%%%%%%%%%%%%
\subsection{Inference-Time Configuration \& Baselines}\label{subsec:inference-setup}

Across both environments we use the E5-Base embedding model \citep{wang2022text} as the shared embedding function $f$. Trajectory chunks span $\winsize = 5$ action-observation steps under the key-value scheme. At inference time, a cascaded retrieval pipeline first retrieves the top 20 candidate trajectory chunks, after which an LTRT reranker selects the final top-1 chunk. The retrieval budget is therefore 1: the working agent conditions on a single retrieved trajectory unit per retrieval call.

\textbf{Model Setups.} We compare three configurations: a vanilla LLM without retrieval, MATM with single-stage dense retrieval only, and MATM with an LTRT reranker (LLM prompts in Appendix~\ref{sec:app:prompts}). 
\textbf{RetrievalPlanner.} Each consumer agent is equipped with a \textit{RetrievalPlanner} LLM that decides, at each interaction step, whether to issue a retrieval call against MATM. This allows agents to call on shared memory selectively rather than on every step, which is important because indiscriminate retrieval can dilute the agent's context with irrelevant guidance. 
\textbf{Metrics.} We evaluate both task performance and efficiency. Task performance is measured by downstream success rate (SR) and efficiency by the number of interaction steps per episode (\# steps). 
To jointly capture both dimensions, we adopt return-paired preference (RPP) \citep{diaz2026rpp}, which compares two configurations task by task, preferring the one that succeeds when the other fails, or succeeds in fewer steps when both succeed. We report each configuration's mean pairwise RPP against all other configurations in the same table, so higher values indicate that a configuration is preferred more often than the alternatives (Appendix~\ref{sec:app:rpp}).
Since consumer agents operate at population scale, unless specified, all reported metrics reflect average performance across consumer models, referred to as consumer population welfare.
All reported numbers are averaged over five runs with different task-model allocation offsets.

\section{Results}

We organize results around five research questions: 
whether MATM augmentation improves the downstream effectiveness and efficiency of consumer agents (\S\ref{subsec:rq1}), whether a learned reranking model can further boost retrieval quality (\S\ref{subsec:rq2}), whether MATM retrieval benefit is exclusive to certain model groups or distributed across the population (\S\ref{subsec:rq3}), whether MATM generalizes across task types (\S\ref{subsec:rq4}), and how consumer population performance scales with memory size (\S\ref{subsec:rq5}).

% \begin{table*}[ht]
% \centering
% \small
% \begin{tabular}{c c c c c c c c}
% \hline
% & \multicolumn{3}{c}{ALFWorld} & \multicolumn{3}{c}{WebArena} \\
% \cline{2-2} \cline{3-5} \cline{6-8}
% \textbf{Method} & \textbf{SR}$\uparrow$ & \textbf{\# steps}$\downarrow$ & \textbf{RPP}$\uparrow$ &\textbf{SR}$\uparrow$ & \textbf{\# steps}$\downarrow$ & \textbf{RPP}$\uparrow$\\
% \hline
% no retrieval & 0 & 0 & 0 & 0 & 0 & 0\\
% \cdashline{1-8}
% single-stage & 0 & 0 & 0 & 0 & 0 & 0\\
% rerank-ffn & 0 & 0 & 0 & 0 & 0 & 0\\
% rerank-lambdamart & 0 & 0 & 0 & 0 & 0 & 0\\
% rerank-svmrank & 0 & 0 & 0 & 0 & 0 & 0\\
% \hline
% \end{tabular}
% \caption{Evaluation of MATM-augmented agents in interactive environments. Success Rate (SR) and number of steps (\# steps) are used for measuring the effectiveness and efficiency. Values are average of the five runs of randomized task-model allocation.}
% \label{tab:interactive-matm}
% \end{table*}

\begin{table*}[t]
\centering
\small
\begin{tabular}{c c c c c c c c}
\hline
& \multicolumn{3}{c}{ALFWorld} & \multicolumn{3}{c}{WebArena} \\
\cline{2-2} \cline{3-5} \cline{6-8}
\textbf{Method} & \textbf{SR}$\uparrow$ & \textbf{\# steps}$\downarrow$ & \textbf{RPP}$\uparrow$ &\textbf{SR}$\uparrow$ & \textbf{\# steps}$\downarrow$ & \textbf{RPP}$\uparrow$\\
\hline
no retrieval & 0.4708 & 11.7664 & -0.1586 & 0.1818 & 21.9886 & -0.0511\\
\cdashline{1-8}
single-stage & 0.5511 & 11.1796 & -0.0451 & \textbf{0.2045} & 20.2614 & 0.0284\\
rerank-FFN & 0.5861 & 11.0336 & -0.0046 & \textbf{0.2045} & \textbf{19.9091} & \textbf{0.0398}\\
rerank-LambdaMART & 0.5715 & 10.8474 & 0.0608 & 0.1818 & 20.4205 & -0.0341\\
rerank-SVMRank & \textbf{0.6431} & \textbf{10.3453} & \textbf{0.1474} & \textbf{0.2045} & 20.2841 & 0.0170\\
\hline
\end{tabular}
\caption{Evaluation of MATM-augmented agents in interactive environments. Success Rate (SR) and number of steps (\# steps) are used for measuring the effectiveness and efficiency.
Return-paired preference (RPP) jointly captures success and efficiency; each value is the mean pairwise RPP of a method against the other four methods in the table (Appendix~\ref{sec:app:rpp}).
Values are average of the five runs of randomized task-model allocation.}
\label{tab:interactive-matm}
\end{table*}

\subsection{MATM-Augmentation improves effectiveness and efficiency}\label{subsec:rq1}
Table~\ref{tab:interactive-matm} summarizes results for ALFWorld and WebArena under no-retrieval and single-stage retrieval from MATM. Across both benchmarks, retrieval from the shared repository consistently improves task outcomes.

On ALFWorld, success rate increases from 47\% to 55\% (+8.0\%p), while 
average steps per episode decrease from 11.77 to 11.18. The mean pairwise RPP rises
from $-0.16$ to $-0.05$, indicating that, in terms of joint success and efficiency, the retrieval-augmented population
is preferred more often than the no-retrieval baseline across the compared methods.
On WebArena, success rate improves from 18\% to 
20\% (+2\%p), with average steps falling from 22.0 to 20.3 and RPP  turning positive at 0.03.

Fewer steps also translate into lower inference cost per solved task: measured directly for single consumer agents, MATM reduces tokens per successful episode by 11--34\%, with nearly identical reductions in dollar cost, despite the added retrieval context (Appendix~\ref{sec:app:token-cost}).

The improvement is more modest than ALFWorld, possibly due to WebArena's longer task horizons and greater sensitivity to early-step errors. Together, these results show that a shared repository of heterogeneous agent trajectories improves consumer population welfare along both effectiveness and efficiency dimensions.

\subsection{Learning to Rank Trajectories further improves MATM participants' welfare}\label{subsec:rq2}

Single-stage retrieval selects trajectories by embedding similarity alone. We next ask whether a learned reranker trained to predict downstream utility can improve over this baseline. We experiment with three reranker configurations: a feed-forward network (FFN), LambdaMART, and SVMRank.

On ALFWorld, all three rerankers improve over single-stage retrieval, and 
SVMRank achieves the strongest results across all metrics: success rate reaches 64.3\% (+9.2\%p over single-stage, +17.2\%p over no-retrieval), average steps fall to 10.35, and RPP rises to 0.15. 
On WebArena, reranker effectiveness is more moderate. FFN matches the success rate of single-stage retrieval at 20.5\% while achieving the lowest step count (19.91) and highest RPP (0.04) among all methods, making it the most effective reranker for that environment. LambdaMART, by contrast, reverts success rate to the no-retrieval level on WebArena, suggesting that the features it relies on are better calibrated to ALFWorld's task structure than WebArena's.
Even in this configuration, average steps still fall below the no-retrieval baseline, so the same fraction of tasks is solved more efficiently; as a result, every retrieval configuration attains a higher mean pairwise RPP than no retrieval across both benchmarks (Table~\ref{tab:interactive-matm}).

\subsection{MATM benefits are distributed across the agent population (Appendix~\ref{sec:app:rq3-full-results})}\label{subsec:rq3}

\begin{figure}[t]
  \centering
  \includegraphics[width=0.95\columnwidth]{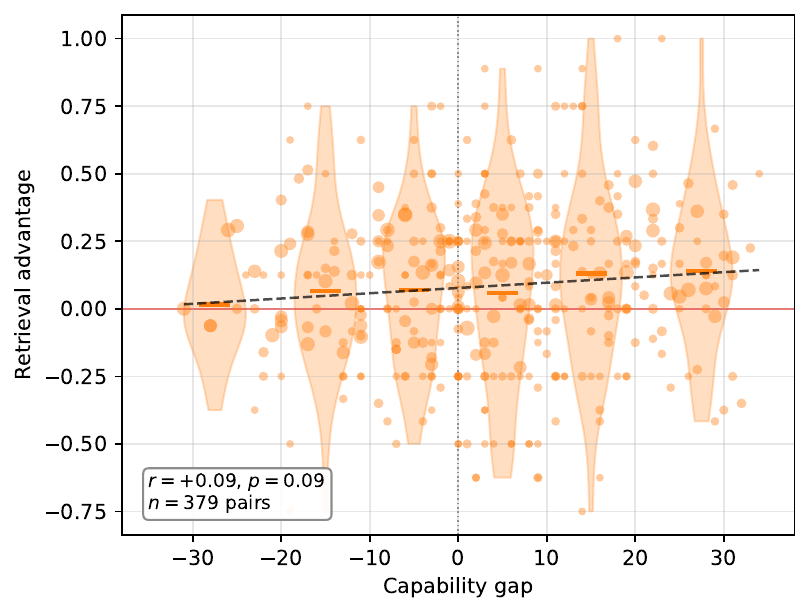}
  \caption{Retrieval Advantage vs. Producer-Consumer Capability Gap on ALFWorld with SVMRank reranking. Each point is one producer--consumer pair.}
  \label{fig:rq3-alfworld-svmrank}
\end{figure}

While the previous sections establish that MATM improves average
consumer welfare, they leave open whether that benefit is concentrated among particular producer-consumer pairings or distributed across the population. To answer this, we measure two quantities for each (producer, consumer) pair: the \textit{retrieval advantage}, defined as the gain in consumer success rate when retrieving from that producer
relative to its no-retrieval baseline, and the \textit{capability gap}, defined as the difference between the two agents' aggregated benchmark scores (AAI~\citep{artificialanalysis2026}).
Higher the capability gap value, the producer is more capable than the consumer.

Figure~\ref{fig:rq3-alfworld-svmrank} shows that consumers benefit from retrieval regardless of whether the producer is weaker, comparable, or stronger than themselves, indicating MATM's value cannot be reduced to a single strong producer. 
Although the correlation between capability gap and retrieval advantage shows a slight positive trend --- suggesting that stronger producers may yield marginally higher benefit --- it remains small and insignificant across both benchmarks, indicating that retrieval utility is not primarily driven by producer-to-consumer competence transfer.
Finally, reranking lifts the entire retrieval advantage distribution, confirming the finding in \S\ref{subsec:rq2}: on ALFWorld, SVMRank roughly doubles the mean retrieval advantage, and the same direction of effect appears on WebArena under FFN reranking (Appendix~\ref{sec:app:rq3-full-results}).

This diversity also holds at the level of retrieved content. Although three producers account for 98.6\% of the ALFWorld index, the remaining 31 producers supply 31.8\% of retrieved trajectories, about 23$\times$ their corpus share, rising to 40.7\% under SVMRank. All 34 producers are retrieved at least once, and the effective number of producers (the exponentiated entropy of the producer distribution) grows from 2.2 in the index to 10.5 in retrieval. On WebArena, whose index is already balanced, retrieval closely tracks corpus share (Appendix~\ref{sec:app:source-diversity}). The skew of the repository therefore does not carry over to what consumers receive.

\subsection{MATM offers cross-task generalization}\label{subsec:rq4}
\begin{table}[t]
\centering
\small
\begin{tabular}{l c c c}
\hline
\textbf{Retrieval Scope} & \textbf{SR} $\uparrow$ & \textbf{\# steps} $\downarrow$ & \textbf{RPP} $\uparrow$ \\
\hline

\textbf{ALFWorld} (274)\\
full & \textbf{0.6460} & \textbf{10.2300} & \textbf{0.0566} \\
same task type & 0.6314 & 10.4635 & 0.0055 \\
cross task type & 0.5985 & 10.7445 & -0.0620 \\

\hline

\textbf{WebArena} (47)\\
full  & \textbf{0.2979} & 19.0426 & \textbf{0.1383} \\
same task type & 0.2766 & 19.1915 & 0.0000 \\
cross task type & 0.1915 & \textbf{19.0000} & -0.1383 \\
\hline
\end{tabular}
\caption{MATM retrieval scope ablation results across three candidate pool restrictions. RPP is the mean pairwise RPP among the three scopes. Number of tasks in parentheses. }
\label{tab:retrieval-scope}
\end{table}

% \begin{table*}[t]
% \centering
% \small
% \begin{tabular}{c c c c c c}
% \hline
% \textbf{Benchmark} & \textbf{Retrieval Scope} & \textbf{\# tasks} & \textbf{SR} $\uparrow$ & \textbf{\# steps} $\downarrow$ & \textbf{RPP} $\uparrow$ \\
% \hline

% \multirow{3}{*}{ALFWorld}
% & full & 274 & \textbf{0.6460} & \textbf{10.2300} & \textbf{0.0566} \\
% & same task & 274 & 0.6314 & 10.4635 & 0.0055 \\
% & cross task & 274 & 0.5985 & 10.7445 & -0.0620 \\

% \hline

% \multirow{3}{*}{WebArena}
% & full & 47 & \textbf{0.2979} & 19.0426 & \textbf{0.1383} \\
% & same task & 47 & 0.2766 & 19.1915 & 0.0000 \\
% & cross task & 47 & 0.1915 & \textbf{19.0000} & -0.1383 \\
% \hline
% \end{tabular}
% \caption{MATM retrieval scope ablation results across three candidate pool restrictions.}
% \label{tab:retrieval-scope}
% \end{table*}

We study how well MATM generalizes across tasks by varying the retrieval scope. We evaluate three conditions: (i) \textit{full} retrieval places no restriction on the candidate pool; (ii) \textit{same-task} retrieval limits candidates to trajectories from the same task type as the query; and (iii) \textit{cross-task} retrieval limits candidates exclusively to trajectories from different task types (Appendix \ref{sec:app:rq4-extra}).

Table~\ref{tab:retrieval-scope} shows that full retrieval achieves the highest SR and RPP in both environments, confirming that unrestricted candidate diversity is beneficial. 
Two findings point to genuine cross-task generalization. First, even under cross-task retrieval, ALFWorld SR reaches 59.9\%, which remains well above the no-retrieval baseline of 47.1\% from  Table~\ref{tab:interactive-matm}. This shows that trajectories from structurally different task types still carry transferable utility. 
Second, the effectiveness and efficiency gap between full and same-task retrieval in both environments suggests that restricting the candidate pool to same-type trajectories is itself a source of degradation, excluding useful candidates that happen to cross task boundaries.
However, for both benchmarks, same-task retrieval outperforms cross-task retrieval, indicating that task-type alignment still carries a meaningful relevance signal.

Although our experiments keep the train and test sets separate, we separately measured an upper bound in which \textit{oracle} trajectories (same task, same environment) from the test set are made retrievable; this lifts SR far above full retrieval. Because such exact matches are rare in practice, we exclude them from our corpus and experiments (Appendix~\ref{sec:app:oracle-ceiling}).

\subsection{MATM scales with memory size}\label{subsec:rq5}

% \begin{figure}[t]
%   \centering
%   \includegraphics[width=\columnwidth]{05-results/rq5/alfworld_scaling_curve.pdf}\\
%   \includegraphics[width=\columnwidth]{05-results/rq5/webarena_scaling_curve.pdf}
%   \caption{MATM memory scaling curves for ALFWorld (top) and WebArena (bottom). Success rate (left axis) and average steps per episode (right axis) as a function of index size. The dotted line marks SR of the no-retrieval baseline. Results are averaged over five runs with different random seeds.}
%   \label{fig:rq5-scaling-curve}
% \end{figure}

\begin{figure}[t]
  \centering
  \includegraphics[width=0.95\columnwidth]{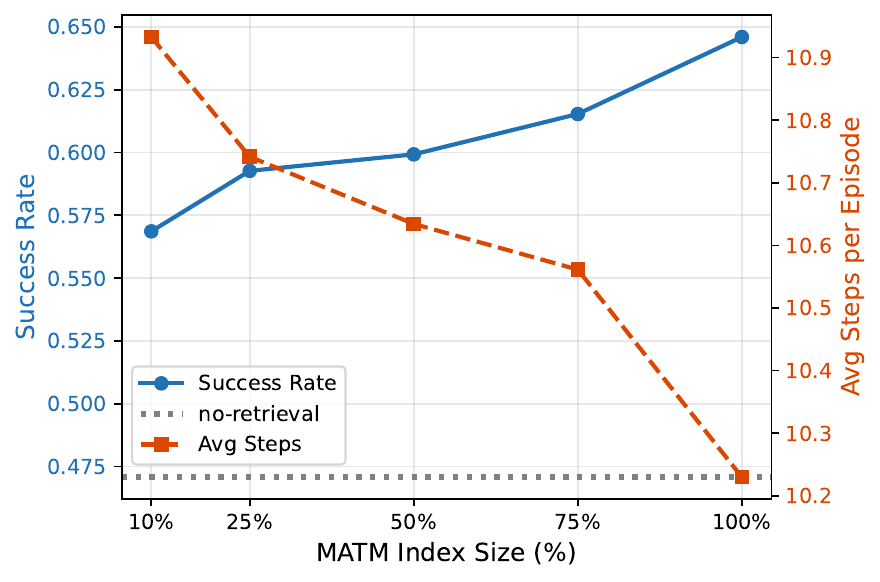}
  \caption{MATM memory scaling curve for ALFWorld. Success rate (left axis) and average steps per episode (right axis) are shown as a function of the MATM index size. The dotted line marks SR of the no-retrieval baseline. Results are averaged over five runs with different random seeds.}
  \label{fig:rq5-scaling-curve-alfworld}
\end{figure}

We study how downstream effectiveness and efficiency change as MATM grows in size. We construct nested memory subsets at 10\%, 25\%, 50\%, 75\%, and 100\% of the full index, with each subset preserving producer model composition and benchmark coverage to isolate the effect of memory size from shifts in data distribution. Each subset is constructed with five different random seeds and results are averaged across runs.

% When full results are in Appendix:
%
Figures~\ref{fig:rq5-scaling-curve-alfworld} shows scaling behavior for ALFWorld. On ALFWorld, both metrics improve monotonically with index size. 
WebArena shows a similar overall trend, though with a non-monotonic pattern in success rate (Appendix \ref{sec:app:rq5-full-results}).
In WebArena, SR shows slight dip at the mid-scale size. This may be because the index is large enough to surface plausible but ultimately unhelpful candidates, yet not large enough to reliably include high-quality matches. At full scale, however, sufficient coverage appears to overcome this noise, restoring and exceeding the gains seen at smaller index sizes.

% Figure~\ref{fig:rq5-scaling-curve} shows the scaling curves for both environments. ALFWorld exhibits monotonic improvement in both success rate and step efficiency with index size, confirming that larger memory consistently benefits the agent.
% On WebArena, step efficiency also decreases monotonically, consistent with ALFWorld. Success rate, however, exhibits a non-monotonic pattern: it dips at the 50\% index before recovering sharply to 20.9\% at full scale, the strongest result across all index sizes and the clearest margin above the no-retrieval baseline of 18.2\%. 
% We hypothesize that at intermediate scales, the index is large enough to surface plausible but ultimately unhelpful trajectories, yet not diverse enough to reliably include high-quality matches. 
% At full scale, sufficient coverage overcomes this noise, restoring and exceeding the gains observed at smaller index sizes.

\section{Discussion}\label{sec:discussion}
Taken together, our results support transactive memory as a viable architecture for population-level agent memory, and several findings sharpen why the population is the right unit of analysis.

% rq1 and rq2
Continued from the results in \S\ref{subsec:rq1} and \S\ref{subsec:rq2}, rerankers' feature importance analysis (Appendix~\ref{sec:app:feature-importance}) suggests that predictive signal extends beyond retrieval similarity to producer metadata such as benchmark scores that represents the producer agents' capability, reframing trajectory selection as producer trust modeling. Because this signal concentrates differently across environments, no single reranker dominates both, underscoring the need for rankers that adapt to task structure rather than a fixed policy.

Moreover, the reduction in interaction steps translates into tangible efficiency gains. Although retrieved context increases input token usage per episode in most settings, MATM reduces both output token usage and monetary cost per successful task (Appendix~\ref{sec:app:token-cost}), demonstrating that shared transactive memory helps agents avoid redundant exploration and improves overall cost efficiency.

% RQ 3
The capability-gap analysis (\S\ref{subsec:rq3}) shows that retrieval benefit is broadly distributed across the population rather than driven by strong-to-weak transfer, which opens per-consumer (group) personalization of the retrieval policy, since the same shared repository may be best exploited differently by agents with different competence profiles.

Likewise, a producer's influence is not necessarily proportional to its contribution volume. On ALFWorld, the 31 minority producers, collectively accounting for only 1.4\% of the index, contribute nearly a third of retrieved trajectories, representing approximately 23$\times$ their corpus share (29$\times$ under SVMRank). All 34 producers are retrieved at least once. In contrast, retrieval on WebArena, where the corpus is more balanced, closely tracks corpus share (Appendix~\ref{sec:app:source-diversity}). These findings suggest that even small contributors can provide substantial value to shared memory, and that producer credit should reflect retrieval frequency and downstream utility rather than corpus share alone.

% RQ4
The retrieval-scope results (\S\ref{subsec:rq4}) show that useful trajectories are not confined within task boundaries. Traditional agent- or task-specific memory would exclude these cross-task candidates by design, and MATM's value lies in making them accessible.

% RQ5
Finally, scaling experiments (\S\ref{subsec:rq5}) demonstrate that value grows with repository size, making incentives for producer contributions an important challenge for such platforms.

\section{Conclusion}\label{sec:conclusion}
We introduced MATM, a shared population-level memory where heterogeneous agents contribute and retrieve trajectories. Retrieval improves effectiveness and efficiency regardless of capability, reranking amplifies these gains, retrieved trajectories generalize across tasks, and benefit scales with memory size. This positions shared artifact storage as a promising substrate for collective, continual lifelong intelligence among distributed agents.

\section*{Limitations and Future Work}
Our experiments cover two interactive benchmarks (ALFWorld and WebArena) and a 34-model consumer population. While no single study can cover every environment or every model in a rapidly evolving landscape, this restricts the scope of our empirical claims. 
Our experiments and rerankers are also trained and evaluated within the same benchmark, so cross-benchmark reranker generalization remains untested.
Also, the LTR dataset uses sparse rank sampling at positions $\mathcal{I} = \{1, 5, 10, 15, 20\}$, which does not fully cover the label distribution at all rank positions. 
This was a practical choice given experiment budget, and we found it to yield strong LTR learning performance with a favorable cost-quality tradeoff.
MATM also indexes only successful trajectories, since its key-value scheme stores successful continuations as guidance; leveraging failed trajectories as negative or contrastive signal is a natural extension.
Finally, our work focuses entirely on the consumer side of MATM.
Because MATM is fundamentally a two-sided market, evaluating producer-side welfare is equally important. 
Future work could draw on attribution fairness in RAG \citep{kim2025fairrag} or marketplace evaluation frameworks \citep{kim2026evaluation} to address this gap. 
Relatedly, the current framework does not account for adversarial producers who may contribute malicious trajectories, potentially placing consumer agents at risk.

\section*{Acknowledgments}
This work is supported by NSF grant 2402874. To Eun Kim is supported by the CMU LTI Presidential Fellowship, endowed by the Asian Student Association, during the 2026--2027 academic year. Any opinions, findings and conclusions or recommendations expressed in this material are those of the authors and do not necessarily reflect those of the sponsors.

% \section*{Author Contributions}
% \begin{itemize}
%     \item To Eun Kim:\\
%     Initial project ideation, implementation of MATM workflow, Learning-to-Rank rerankers, and LTRT dataset design, writing, review rebuttals.
%     \item Xuhong He:\\
%     LTRT dataset collection, trajectory reranker training and experiments on ALFWorld and WebArena
%     \item Dishank Jain:\\
%     LTRT dataset collection, trajectory reranker training and experiments on ALFWorld and WebArena
%     \item Ambuj Agrawal :\\
%     Preliminary experiments on cross-task generalization
%     \item Negar Arabzadeh:\\
%     Initial project ideation
%     \item Fernando Diaz:\\
%     Initial project ideation, writing, project supervision
% \end{itemize}

\bibliography{XX-references.bib}

\appendix
\section{Unabridged Related Work}\label{sec:app:related-work}

\subsection{Memory-Augmented Agents}
Retrieval-augmented generation (RAG) \citep{lewis2020retrieval}, an instance of Retrieval-Enhanced Machine Learning \citep{zamani:reml, kim2024reml}, enhances language models by conditioning generation on retrieved external context, most commonly human-authored documents such as web pages or knowledge bases \citep{fan2024survey}. Recent extensions of this paradigm treat an agent's own interaction history as retrievable context, giving rise to memory-augmented generation, where past conversations or execution traces are indexed and reused to guide future behavior \citep{shinn2023reflexion, majumder2024clin, zheng2024synapse}.

MATM fits within this retrieval- and memory-augmented paradigm but differs in both content and scope: instead of retrieving web documents or single-agent's local history, MATM indexes agent-generated trajectories and treats memory as a population-level resource shared across agents, where agents can contribute to and retrieve from the shared repository.

\subsection{Reuse of Agent Artifacts}
Modern reasoning agents, particularly those employing high inference-time scaling, generate rich intermediate artifacts during problem solving. These include low-level action-observation trajectories and thinking traces, as well as higher-level plans, strategies, workflows, and reusable code which are anologous to the notion of options in reinforcement learning \citep{Garcia19compressionMacro, veeriah2021discovery}. These \textit{agent artifacts} can be leveraged to improve inference efficiency, generalization, and continual adaptation.

At the trajectory level, prior work has explored reusing reasoning or action-observation trajectories as in-context guidance. 
Buffer of Thoughts \citep{yang2024buffer} maintains and retrieves reasoning templates to guide new problem instances, while Retrieval of Thought \citep{ahmed2025retrieval} constructs thought templates on the fly by retrieving prior reasoning trajectories. 
For action-observation trajectories, \citet{zheng2024synapse} and \citet{zhao2024expel} reuse environment interaction trajectories as in-context examples to improve downstream decision-making. 

Beyond the trajectory level, several works extract and reuse more abstract artifacts such as plans, strategies, workflows, and skills. 
CLIN \citep{majumder2024clin} stores textual causal abstractions to support continual improvement. Agent Workflow Memory \citep{wang2025agentworkflow} distills reusable workflows from web interaction trajectories, and MaestroMotif \citep{klissarov2025maestromotif} induces reusable skills via reinforcement learning. ReasoningBank \citep{ouyang2025reasoningbank} retrieves strategy-level reasoning patterns to guide problem solving.  \citet{arabzadeh2026thinkingtrace} transform math reasoning thinking trajectories into higher level structures and use them as retrievable objects.  
Applied to a programming domain, \citet{wang2025inducing} enable agents to induce, verify, and reuse program-based skills on the fly in web-based tasks, while Voyager \citep{wang2024voyager} maintains a growing library of executable code for open-ended task execution. In a concurrent work, SkillNet \citep{liang2026skillnet} assembles a collection of skills contributed by multiple agents, framing skill accumulation as a system-design problem. 
AgentRxiv \citep{schmidgall2025agentrxiv} pursues a similar direction for autonomous research, providing a shared preprint server through which agent laboratories upload and retrieve research reports. Its experiments involve laboratories with a common research objective, leaving reuse in open ecosystems of heterogeneous agents untested.
Collectively, these works frame artifact reuse as a mechanism for accumulating reusable competence over time.

Agent artifacts can also serve as supervision signals for model distillation. SuperCorrect \citep{yang2025supercorrect} extracts thought templates from a teacher model to guide smaller models during reasoning. Related approaches similarly distill structured reasoning artifacts to transfer competence across models \citep{li2025naturalthoughts, kang2025distilling}. 
In these settings, artifacts function not only as inference-time memory but also as compressed representations of reasoning expertise.

While these systems reuse different types of agent artifacts, such artifacts are typically reused only by the same or homogeneous agent(s) that produced them, with less consideration of emerging society of agents \citep{liDoesSocializationEmerge2026, wangSkillOrchestraLearningRoute2026}.
As a result, valuable experience remains isolated, and newly instantiated agents repeatedly rediscover solutions that already existed elsewhere in the other systems.
This gap is increasingly recognized at the level of agent populations. \citet{jordan2025collectivist} argues that intelligence is fundamentally collective and that algorithm design should target social welfare rather than individual performance, while \citet{hammond2025multi} anticipate the imminent deployment of heterogeneous agent populations and identify information asymmetry as a central challenge. \citet{tomavsev2025distributional} envision shared retrieval repositories of agent experience whose value depends on separating novel, high-utility contributions from redundant, low-utility ones, noting that information-value estimators are needed to do so. MATM provides exactly this substrate: a shared, population-level repository with retrieval and ranking mechanisms over agent trajectories.

In contrast, MATM proposes a \textit{population-level} artifact repository. Rather than treating artifacts as private, per-agent memory, we model them as shared, structured resources that heterogeneous agents can both contribute to and retrieve from. This shifts artifact reuse from an individual optimization mechanism to a collective knowledge infrastructure, enabling continual learning and cross-agent transfer, reducing redundant exploration, and supporting cumulative capability growth at the ecosystem level.
The most closely related concurrent work, SkillNet \citep{liang2026skillnet}, also assembles a collection of skills across heterogeneous agents. However, it primarily addresses system design and does not evaluate the benefit of retrieval for consumer agents in a population setting, nor does it provide in-depth analysis of artifact repository search.

\section{Return-Paired Preference \citep{diaz2026rpp}}\label{sec:app:rpp}
Given a set of $n$ tasks and two agents---a control or baseline agent $\agent$ and a treatment agent $\agent'$---we have a set of $n$ trajectories for each agent.  For a task, we say that $\agent'$ is preferred to $\agent$ if $\agent'$ is successful and $\agent$ is not or, if both are successful, $\agent'$ reaches the success in fewer steps; similarly, $\agent$ is preferred to $\agent'$ if $\agent$ is successful while $\agent'$ is not or if it faster to success if both are successful.  In all other situations, we say that there is no preference between $\agent'$ and $\agent$ for that task.  If the value of the preference is $1$ when $\agent'$ is preferred, $-1$ when $\agent$ is preferred, and $0$ if there is no preference, then the pairwise return-paired preference $\mathrm{RPP}(\agent', \agent) \in [-1, 1]$ is the \textbf{mean} preference value over all $n$ tasks. Note that $\mathrm{RPP}(\agent', \agent) = -\mathrm{RPP}(\agent, \agent')$.

Within each results table, we compare a set of configurations $\mathcal{M}$ (e.g., the five methods in Table~\ref{tab:interactive-matm}, or the three retrieval scopes in Table~\ref{tab:retrieval-scope}) evaluated on the same tasks. The RPP we report for a configuration $\agent' \in \mathcal{M}$ is its \textbf{mean pairwise} RPP against every other configuration in that set,
\[
\overline{\mathrm{RPP}}(\agent') = \frac{1}{|\mathcal{M}| - 1} \sum_{\agent \in \mathcal{M} \setminus \{\agent'\}} \mathrm{RPP}(\agent', \agent).
\]
A positive value means the configuration is preferred over the other configurations more often than not, and a negative value means it is dispreferred more often. By antisymmetry, the mean pairwise RPP values within a table sum to zero, so no configuration (including the no-retrieval baseline) is fixed at zero; values are comparable only within the same table.

\section{Population of LLM Agents}\label{sec:app:population-list}
Table \ref{tab:agent-population-list} shows the list of agents used as a population in each benchmark.
\begin{table}[t]
\centering
\resizebox{\columnwidth}{!}{
\begin{tabular}{l c c}
\hline
\textbf{Model} & \textbf{Producer} & \textbf{Consumer} \\
\hline
trained-seq2seq                    & A   &     \\
openai/gpt-4-turbo                 & W   &     \\
openai/gpt-4-turbo-preview         & W   &     \\
anthropic/claude-3.5-sonnet        & W   &     \\
openai/gpt-oss-20b                 & AW  & AW  \\
openai/gpt-oss-120b                & AW  & AW  \\
openai/gpt-5.4                     & AW  & AW  \\
openai/gpt-5.4-nano                & AW  & AW  \\
anthropic/claude-3-haiku           & AW  & AW  \\
anthropic/claude-sonnet-4.6        & AW  & AW  \\
anthropic/claude-opus-4            & AW  & AW  \\
anthropic/claude-opus-4.6          & AW  & AW  \\
google/gemini-2.5-flash-lite       & AW  & AW  \\
google/gemini-2.5-flash            & AW  & AW  \\
google/gemini-2.5-pro              & AW  & AW  \\
google/gemma-4-31b-it              & AW  & AW  \\
meta-llama/llama-3.1-8b-instruct   & AW  & AW  \\
meta-llama/llama-3.2-3b-instruct   & AW  & AW  \\
meta-llama/llama-3.3-70b-instruct  & AW  & AW  \\
meta-llama/llama-4-maverick        & AW  & AW  \\
deepseek/deepseek-r1               & AW  & AW  \\
deepseek/deepseek-r1-0528          & AW  & AW  \\
deepseek/deepseek-chat-v3.1        & AW  & AW  \\
deepseek/deepseek-v3.2             & AW  & AW  \\
qwen/qwen3-32b                     & AW  & AW  \\
qwen/qwen3-235b-a22b               & AW  & AW  \\
qwen/qwen3-max                     & AW  & AW  \\
qwen/qwen3.5-flash-02-23           & AW  & AW  \\
x-ai/grok-3-mini                   & AW  & AW  \\
x-ai/grok-4-fast                   & AW  & AW  \\
x-ai/grok-4.1-fast                 & AW  & AW  \\
x-ai/grok-4.20                     & AW  & AW  \\
z-ai/glm-4.5                       & AW  & AW  \\
z-ai/glm-5                         & AW  & AW  \\
minimax/minimax-m1                 & AW  & AW  \\
minimax/minimax-m2                 & AW  & AW  \\
mistralai/ministral-14b-2512       & AW  & AW  \\
mistralai/mistral-large-2512       & AW  & AW  \\
\hline
\end{tabular}
}
\caption{Producer and consumer agents across ALFWorld (A) and WebArena (W). AW indicates the model serves in that role for both environments.}
\label{tab:agent-population-list}
\end{table}

\section{ALFWorld Description}\label{sec:app:alfworld-split}
ALFWorld \citep{shridhar2021alfworld} contains interactive TextWorld environments \citep{cote18textworld} that parallel embodied worlds in the ALFRED dataset \citep{ALFRED20}. The aligned environments allow agents to reason and learn high-level policies in an abstract space before solving embodied tasks through low-level actuation. ALFWorld translates complex household tasks such as finding, cleaning, heating, or placing objects into textual observations and actions, allowing researchers to train and evaluate agents using natural language rather than raw visual input. The dataset consists of 3553 tasks for training  and a heldout test set of 274 tasks. The tasks are grouped into 6 task types: Pick \& Place, Examine in Light, Clean \& Place, Heat \& Place, Cool \& Place, Pick Two \& Place. Within each task category there is significant variation: the embodied environment includes 120
rooms (30 kitchens, 30 bedrooms, 30 bathrooms, 30 living rooms), each dynamically populated with a set of portable objects (e.g., apple, mug), and static receptacles (e.g., microwave, fridge). For interaction, TextWorld environments allow 9 high-level actions such as 'open', 'heat', etc. For our experiments, we use the 3553 training episodes for populating MATM memory, and a sampled representative subset of 355 episodes for generating data for training LTRT rerankers. We use the heldout 274 episodes as the test set.

\section{WebArena Description}\label{sec:app:webarena-split}
\begin{figure}
    \centering
    \includegraphics[width=\linewidth]{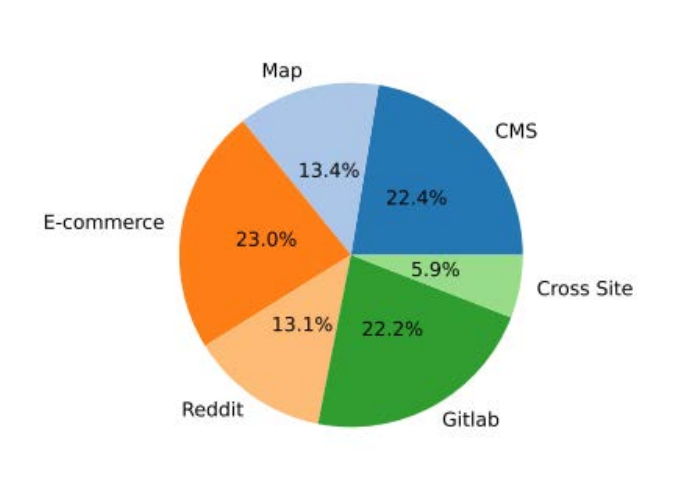}
    \caption{The intent distribution across different websites for WebArena}
    \label{fig:webarena-intents}
\end{figure}

WebArena \citep{zhou2024webarena} is a standalone, self-hostable web environment for building autonomous agents. WebArena creates websites from five popular categories (Ecommerce platforms, Social Forums, Maps, Content Management Systems and Collaborative Development Platforms for software development) with functionality and data mimicking their real-world equivalents. The dataset consists of 812 examples consisting of high-level natural language instructions that require interaction with the WebArena environment to solve. The dataset was created by curating realistic intents to carry out complex and creative tasks within WebArena. Annotators were guided to spend a few minutes exploring the websites to familiarize themselves with the websites’ content and functionalities. Then the annotators are tasked with intent formulation. At the end, 241 intents were curated and 812 tasks were created with different instantiations of these intents. Figure \ref{fig:webarena-intents} shows the distribution of intents accross different sites.

For our experiments, the test set comprised of 88 tasks sampled from the total 812 tasks while maintaining the distribution of intents. We treated the leftover 724 tasks as training data for populating MATM memory, and used a subset of 58 episodes for generating LTRT training data. In the end, our test set consisted of 88 tasks.

\section{Task Allocation Function}\label{sec:app:allocation_function}
To ensure that every model in the population produces trajectories across all task categories, we employ a \textit{task-type-aware stratified round-robin} assignment with an offset.
Given a partition $\instanceSpace_p$, questions are first grouped into buckets by task type (e.g., Algebra, Geometry, Precalculus for mathematical problem solving, or task categories for interactive benchmarks), and each bucket is sorted in a deterministic order.
Within each bucket, questions are assigned to agents by cycling through the ordered population $\agent_1, \agent_2, \ldots, \agent_N$ in round-robin fashion, starting at an offset $o \in \{0,\ldots,N{-}1\}$.
The offset shifts the starting agent but does not change the bucket composition, ensuring that different offset values produce complementary assignments across agents.

\section{MATM Index Statistics}\label{sec:app:index-stat}
Table \ref{tab:index_stat_all} shows the size of MATM index across environments.
\begin{table}[ht]
\centering
\scriptsize

\begin{subtable}[t]{0.49\columnwidth}
\centering
\resizebox{\columnwidth}{!}{
\begin{tabular}{l c c}
\hline
\textbf{Producer Agents} & \textbf{\# chunks} & \textbf{\%} \\
\hline
trained-seq2seq                    & 67,570 & 77.82\% \\
Qwen3-32B                          & 11,264 & 12.97\% \\
GPT-OSS 20B                        & 6,781  & 7.81\%  \\
DeepSeek-V3.2                      & 85     & 0.10\%  \\
MiniMax-M1                         & 68     & 0.08\%  \\
Qwen3.5-Flash                      & 68     & 0.08\%  \\
Claude Sonnet 4.6                  & 66     & 0.08\%  \\
GLM-4.5                            & 63     & 0.07\%  \\
DeepSeek-R1                        & 59     & 0.07\%  \\
Claude Opus 4                      & 54     & 0.06\%  \\
MiniMax-M2                         & 54     & 0.06\%  \\
Gemini 2.5 Flash                   & 52     & 0.06\%  \\
DeepSeek-Chat-V3.1                 & 52     & 0.06\%  \\
Qwen3-Max                          & 50     & 0.06\%  \\
Grok-4-Fast                        & 45     & 0.05\%  \\
Gemini 2.5 Pro                     & 41     & 0.05\%  \\
Grok-4.1-Fast                      & 39     & 0.04\%  \\
Gemini 2.5 Flash Lite              & 36     & 0.04\%  \\
GLM-5                              & 35     & 0.04\%  \\
Mistral Large 2512                 & 34     & 0.04\%  \\
Gemma-4-31B                        & 33     & 0.04\%  \\
Claude Opus 4.6                    & 32     & 0.04\%  \\
Ministral-14B                      & 30     & 0.03\%  \\
DeepSeek-R1-0528                   & 30     & 0.03\%  \\
Qwen3-235B-A22B                    & 29     & 0.03\%  \\
Grok-4.20                          & 27     & 0.03\%  \\
Llama-3.3-70B                      & 24     & 0.03\%  \\
GPT-OSS 120B                       & 23     & 0.03\%  \\
Llama-4-Maverick                   & 23     & 0.03\%  \\
Grok-3-Mini                        & 21     & 0.02\%  \\
Claude 3 Haiku                     & 17     & 0.02\%  \\
GPT-5.4                            & 12     & 0.01\%  \\
GPT-5.4-Nano                       & 9      & 0.01\%  \\
Llama-3.1-8B                       & 7      & 0.01\%  \\
\hline
\textbf{Total (34)}                     & \textbf{86,833} & \textbf{100\%} \\
\hline
\end{tabular}
}
\caption{ALFWorld}
\end{subtable}
% \hfill
\begin{subtable}[t]{0.49\columnwidth}
\centering
\resizebox{\columnwidth}{!}{
\begin{tabular}{l c c}
\hline
\textbf{Producer Agents} & \textbf{\# chunks} & (\%) \\
\hline
GPT-4-Turbo                        & 2,222  & 11.05\% \\
GPT-4-Turbo-Preview                & 2,054  & 10.22\% \\
Claude 3.5 Sonnet                  & 1,673  & 8.32\%  \\
Qwen3-32B                          & 1,500  & 7.46\%  \\
GPT-OSS 20B                        & 1,098  & 5.46\%  \\
DeepSeek-R1-0528                   & 1,040  & 5.17\%  \\
Claude Opus 4                      & 982    & 4.89\%  \\
MiniMax-M1                         & 804    & 4.00\%  \\
Claude Sonnet 4.6                  & 801    & 3.98\%  \\
Claude Opus 4.6                    & 725    & 3.61\%  \\
DeepSeek-R1                        & 674    & 3.35\%  \\
Gemini 2.5 Flash                   & 673    & 3.35\%  \\
GLM-5                              & 577    & 2.87\%  \\
Llama-4-Maverick                   & 442    & 2.20\%  \\
Grok-4.1-Fast                      & 437    & 2.17\%  \\
Grok-4.20                          & 431    & 2.14\%  \\
Qwen3.5-Flash                      & 411    & 2.04\%  \\
DeepSeek-V3.2                      & 404    & 2.01\%  \\
GPT-5.4                            & 388    & 1.93\%  \\
GLM-4.5                            & 342    & 1.70\%  \\
GPT-5.4-Nano                       & 306    & 1.52\%  \\
MiniMax-M2                         & 289    & 1.44\%  \\
Grok-4-Fast                        & 262    & 1.30\%  \\
Qwen3-Max                          & 246    & 1.22\%  \\
Qwen3-235B-A22B                    & 224    & 1.11\%  \\
Gemini 2.5 Pro                     & 217    & 1.08\%  \\
Grok-3-Mini                        & 203    & 1.01\%  \\
GPT-OSS 120B                       & 158    & 0.79\%  \\
DeepSeek-Chat-V3.1                 & 139    & 0.69\%  \\
Gemma-4-31B                        & 118    & 0.59\%  \\
Mistral Large 2512                 & 116    & 0.58\%  \\
GPT-4-1106-Preview                 & 89     & 0.44\%  \\
Llama-3.1-8B                       & 22     & 0.11\%  \\
Llama-3.3-70B                      & 20     & 0.10\%  \\
Gemini 2.5 Flash Lite              & 11     & 0.05\%  \\
Claude 3 Haiku                     & 4      & 0.02\%  \\
\hline
\textbf{Total (36)}                     & \textbf{20,102} & \textbf{100\%} \\
\hline
\end{tabular}
}
\caption{WebArena}
\end{subtable}

\caption{MATM index statistics across benchmarks.}
\label{tab:index_stat_all}
\end{table}

\section{Retrieved-Source Diversity}\label{sec:app:source-diversity}
\begin{table}[t]
\centering
\small
\resizebox{\columnwidth}{!}{
\begin{tabular}{l c c c c}
\hline
\textbf{Producer} & \textbf{Index (\%)} & \textbf{Retrieved (\%)} & \textbf{Retrieved (\%)} & \textbf{Pooled /} \\
 & & \textbf{pooled} & \textbf{best reranker} & \textbf{Index} \\
\hline
\multicolumn{5}{l}{\textbf{ALFWorld} (best reranker: SVMRank)} \\
\hline
trained-seq2seq & 77.82 & 35.3 & 25.6 & 0.45$\times$ \\
Qwen3-32B & 12.97 & 22.1 & 23.7 & 1.70$\times$ \\
GPT-OSS 20B & 7.81 & 10.8 & 9.9 & 1.38$\times$ \\
\textbf{Other 31 producers} & \textbf{1.40} & \textbf{31.8} & \textbf{40.7} & \textbf{22.68$\times$} \\
\cdashline{1-5}
\# producers retrieved & -- & 34 & 32 & \\
Effective \# producers & 2.2 & 10.5 & 11.0 & \\
\hline\hline
\multicolumn{5}{l}{\textbf{WebArena} (best reranker: FFN)} \\
\hline
GPT-4-Turbo & 11.05 & 3.2 & 3.1 & 0.29$\times$ \\
GPT-4-Turbo-Preview & 10.22 & 6.8 & 9.9 & 0.67$\times$ \\
Claude 3.5 Sonnet & 8.32 & 9.2 & 4.9 & 1.10$\times$ \\
Qwen3-32B & 7.46 & 13.3 & 17.9 & 1.79$\times$ \\
GPT-OSS 20B & 5.46 & 10.0 & 5.6 & 1.83$\times$ \\
\textbf{Other 31 producers} & \textbf{57.48} & \textbf{57.5} & \textbf{58.6} & \textbf{1.00$\times$} \\
\cdashline{1-5}
\# producers retrieved & -- & 31 & 25 & \\
Effective \# producers & 23.4 & 20.4 & 18.4 & \\
\hline
\end{tabular}
}
\caption{Producer shares of the MATM index versus shares of retrieved trajectories. \textit{Pooled} aggregates retrievals over all consumers and over single-stage retrieval and the three rerankers. The effective number of producers is $\exp(H)$, where $H$ is the entropy of the producer distribution.}
\label{tab:source-diversity}
\end{table}

The ALFWorld index is dominated by three producers (Table~\ref{tab:index_stat_all}), but what a consumer agent conditions on is determined by which trajectories are actually retrieved. Table~\ref{tab:source-diversity} compares each producer's share of the index with its share of retrieved trajectories.

On ALFWorld, the 31 minority producers hold only 1.4\% of the index yet supply 31.8\% of retrievals, about 23$\times$ their index share, rising to 40.7\% under SVMRank. All 34 producers are retrieved, and the effective number of producers grows from 2.2 in the index to 10.5 in retrieval. On WebArena, the index is already balanced (no producer exceeds 11.1\% of chunks), and retrieval closely tracks it. \textbf{Hence, the dominant corpus sources do not dominate what consumers actually receive.}

\section{Incremental Construction of MATM \& LTRT Dataset}\label{app:sec:data_construction}
\begin{algorithm*}[t]
\caption{Incremental MATM \& LTRT Dataset Construction}
\label{alg:matm-construction}
\begin{algorithmic}[1]
\Require Agent population $\mathcal{\agent}$, question partitions $\instanceSpace_1, \dots, \instanceSpace_P$, pre-warmed index $\corpus_0$, retrieval depth $K$, rank positions $\mathcal{I} \subseteq \{1,\dots,K\}$, branching points per question $T$, allocation $\sigma$, evaluator $\textsc{Eval}$, quality threshold $\metricThreshold$, embedding function $f$
\Ensure Updated MATM index $\corpus_P$ and LTRT dataset $\ltrtDataset$
\State $\ltrtDataset \gets \emptyset$ \Comment{LTRT Dataset}
\For{$p = 1, \dots, P$} \Comment{Process each partition}
    \State $\mathcal{B}_p \gets \emptyset$ \Comment{Trajectory buffer}
    \For{each question $\instance \in \instanceSpace_p$}
        \State $\agent_n \gets \sigma(\instance, \mathcal{\agent}, p)$ \Comment{Assign agent}
        \State $(\trajectory_{\mathrm{base}}, \output_{\mathrm{base}}) \gets \agent_n(\instance)$ \Comment{Baseline trajectory without retrieval}
        \State $s_{\mathrm{base}} \gets \textsc{Eval}(\output_{\mathrm{base}}, \instance)$ \Comment{Reference score for marginal utility}
        \If{$s_{\mathrm{base}} \geq \metricThreshold$}
            \State $\mathcal{B}_p \gets \mathcal{B}_p \cup \{ \trajectory_{\mathrm{base}} \}$
        \EndIf
        \State Sample $\{t_1, \dots, t_T\} \subseteq \{1, \dots, |\trajectory_{\mathrm{base}}|\}$ uniformly at random \Comment{Branching points}
        \For{$t \in \{t_1, \dots, t_T\}$} \Comment{Roll-in to step $t$ of $\trajectory_{\mathrm{base}}$}
            \State $h_t \gets (\tau_1, \dots, \tau_t)$ from $\trajectory_{\mathrm{base}}$
            \State $\query_t, (\doc_t^{(1)},\!\dots,\!\doc_t^{(K)}) \gets \textsc{Retrieve}(\instance, h_t, \corpus_{p-1}, K)$
            \For{$j \in \mathcal{I}$} \Comment{Roll-out: one-shot augmented generation per rank}
                \State $(\trajectory_t^{(j)}, \output_t^{(j)}) \gets \agent_n(\instance \mid h_t, \doc_t^{(j)})$
                \State $s_t^{(j)} \gets \textsc{Eval}(\output_t^{(j)}, \instance)$
                \If{$s_t^{(j)} \geq \metricThreshold$}
                    \State $\mathcal{B}_p \gets \mathcal{B}_p \cup \{ \trajectory_t^{(j)} \}$ \Comment{Add successful trajectory}
                \EndIf
                \State $\ltrtDataset \gets \ltrtDataset \cup \{ (\query_t, \doc_t^{(j)}, s_t^{(j)} - s_{\mathrm{base}}) \}$ \Comment{Marginal utility label}
            \EndFor
        \EndFor
    \EndFor
    \State $\corpus_p \gets \textsc{IndexUpdate}(\corpus_{p-1},\; \mathcal{B}_p,\; f)$ \Comment{Chunk, embed, add}
\EndFor
\end{algorithmic}
\end{algorithm*}
Algorithm \ref{alg:matm-construction} describes formal procedure of incremental construction of MATM index and LTRT dataset for trajectory reranker training.

\section{Learning-To-Rank Features}\label{sec:app:ltr-features}
Table \ref{tab:ltr-features-full-list} shows the complete list of learning-to-rank features used across environments.
\begin{table}[ht]
\centering
\scriptsize
\setlength{\tabcolsep}{4pt}
\resizebox{\columnwidth}{!}{%
\begin{tabular}{ll}
\hline
\textbf{Category} & \textbf{Features (\# features: 44)} \\
\hline
Producer Agent Info (\#: 13)
  & agent ID \\
  & context-window \\
  & agent benchmark scores (11 features):\\
  & \quad Artificial Analysis Intelligence (AAI) Index \\
  & \quad GDPval-AA \\
  & \quad $\tau^2$-Bench Telecom \\
  & \quad Terminal-Bench Hard \\
  & \quad SciCode \\
  & \quad AA-LCR \\
  & \quad AA-Omniscience Accuracy \\
  & \quad IFBench \\
  & \quad Humanity's Last Exam (HLE) \\
  & \quad GPQA-Diamond \\
  & \quad CritPt \\
\hline
Consumer Agent Info (\#: 1)
  & agent ID \\
\hline
1\textsuperscript{st} Stage Retrieval (\#: 1)
  % & 1\textsuperscript{st} stage retrieval rank \\
  & 1\textsuperscript{st} stage retrieval score \\
\hline
Query Features (\#: 2)
  & query length \\
  & current step number \\
\hline
Trajectory Features (\#: 4)
  & retrieved chunk length \\
  & number of steps in trajectory \\
  & success flag \\
  & trajectory length \\
\hline
Query--Trajectory Interaction Features (\#: 23)
  & unigram text tfidf cosine similarity \\
  & unigram goal tfidf cosine similarity \\
  & unigram state tfidf cosine similarity \\
  & unigram context tfidf cosine similarity \\
  & bigram text tfidf cosine similarity \\
  & bigram goal tfidf cosine similarity \\
  & bigram state tfidf cosine similarity \\
  & bigram context tfidf cosine similarity \\
  & text overlap ratio \\
  & goal overlap ratio \\
  & state overlap ratio \\
  & context overlap ratio \\
  & text jaccard similarity \\
  & goal jaccard similarity \\
  & state jaccard similarity \\
  & context jaccard similarity \\
  & text embedding similarity \\
  & goal embedding similarity \\
  & state embedding similarity \\
  & context embedding similarity \\
  & task match \\
  & task variation match \\
  & step number difference \\
\hline
\end{tabular}
}
\caption{Features used for Learning-To-Rank Trajectories (LTRT).}
\label{tab:ltr-features-full-list}
\end{table}

\section{Feature Importance Test of the Trained Learning-To-Rank Trajectories (LTRT) Model}\label{sec:app:feature-importance}
\begin{table*}[ht]
\centering
\small
\begin{tabular}{c c c c c}
\hline
\textbf{Benchmark} & \textbf{Rank} & \textbf{Feature Name} & \textbf{Feature Category} & \textbf{Feature Importance} \\
\hline
AlfWorld & 1 &  SciCode score & Producer Agent Info & 0.0596 \\
AlfWorld & 2 & number of steps in trajectory & Trajectory Features & -0.0514\\
AlfWorld & 3 &  AAI score & Producer Agent Info & 0.0368 \\
AlfWorld & 4 &  HLE score & Producer Agent Info & -0.0233 \\
AlfWorld & 5 &  GPQA-Diamond score & Producer Agent Info & -0.0230 \\
AlfWorld & 6 &  state overlap ratio & Query--Trajectory Interaction & 0.0188 \\
AlfWorld & 7 &  GDPval-AA score & Producer Agent Info & 0.0185 \\
AlfWorld & 8 &  $\tau^2$-Bench Telecom score & Producer Agent Info & -0.0168 \\
AlfWorld & 9 & retrieved chunk length & Trajectory Features & 0.0156 \\
AlfWorld & 10 &  IFBench score & Producer Agent Info & -0.0155 \\
\hline
WebArena & 1 &  AA-LCR score & Producer Agent Info & 0.0069 \\
WebArena & 2 &  IFBench score & Producer Agent Info & 0.0063 \\
WebArena & 3 & number steps in trajectory & Trajectory Features & 0.0044 \\
WebArena & 4 & state embedding similarity & Query--Trajectory Interaction & 0.0036 \\
WebArena & 5 & goal overlap ratio & Query--Trajectory Interaction & 0.0032 \\
WebArena & 6 & GPQA-Diamond score & Producer Agent Info & 0.0028 \\
WebArena & 7 & retrieved chunk length & Trajectory Features & 0.0024 \\
WebArena & 8 &  CritPt score & Producer Agent Info & 0.0021 \\
WebArena & 9 & current step number & Query Features & 0.0020 \\
WebArena & 10 & bigram text tfidf cosine similarity & Query--Trajectory Interaction & 0.0019 \\
\hline
\end{tabular}
\caption{Top feature rankings for LTRT model on each benchmark. SVMRank for ALFWorld and FFN for WebArena. For SVMRank, we report the feature importance as the weight learnt for SVMRank. For FFN, we compute feature importance by removing the feature and measuring the drop in NDCG@10 score while training the LTRT reranker.}
\label{tab:feature-importance}
\end{table*}

Table \ref{tab:feature-importance} shows the top ten most important features for both benchmarks.

\section{Extended Results of Section \S\ref{subsec:rq3}}\label{sec:app:rq3-full-results}

\subsection{Formalism}
We formalize the analysis of \S\ref{subsec:rq3} as follows. 
With some abuse of notation, let
$\mathcal{P}$ denote the set of producer agents and $\mathcal{C}$ the set of consumer agents, with $\mathcal{X}$ the set of evaluation tasks. For a consumer $c \in \mathcal{C}$, let $\mu_0(c) \in [0, 1]$ denote its average final score on $\mathcal{X}$ without retrieval, and let $\mu_r(p, c) \in [0, 1]$ denote its average final score on episodes where producer $p \in \mathcal{P}$ appeared
among the retrieved source models. The \textit{retrieval advantage} of
the pair $(p, c)$ is
\[
  \mu_\alpha(p, c) = \mu_r(p, c) - \mu_0(c),
\]
the gain in consumer success rate attributable to retrieving from
producer $p$ relative to that consumer's own no-retrieval baseline.

Each agent $i \in \mathcal{P} \cup \mathcal{C}$ has a standalone
capability $\kappa(i)$, measured by its aggregated Artificial Analysis
Intelligence Index score~\citep{artificialanalysis2026}. The
\textit{capability gap} of a producer-consumer pair is
\[
  \kappa_\alpha(p, c) = \kappa(p) - \kappa(c),
\]
which is positive when the producer is stronger than the consumer in
standalone capability, zero when they are matched, and negative when
the producer is weaker.

\begin{figure}[h]
  \centering
  \includegraphics[width=\columnwidth]{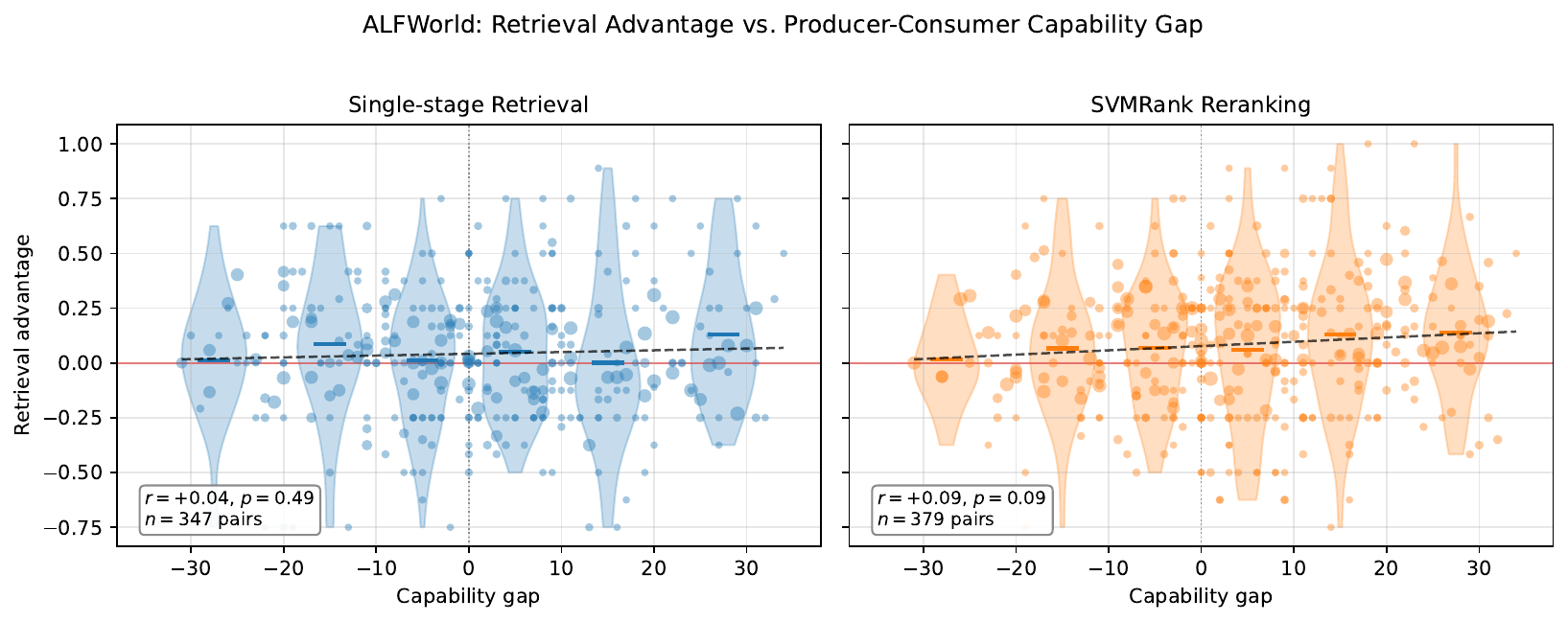}\\
  \includegraphics[width=\columnwidth]{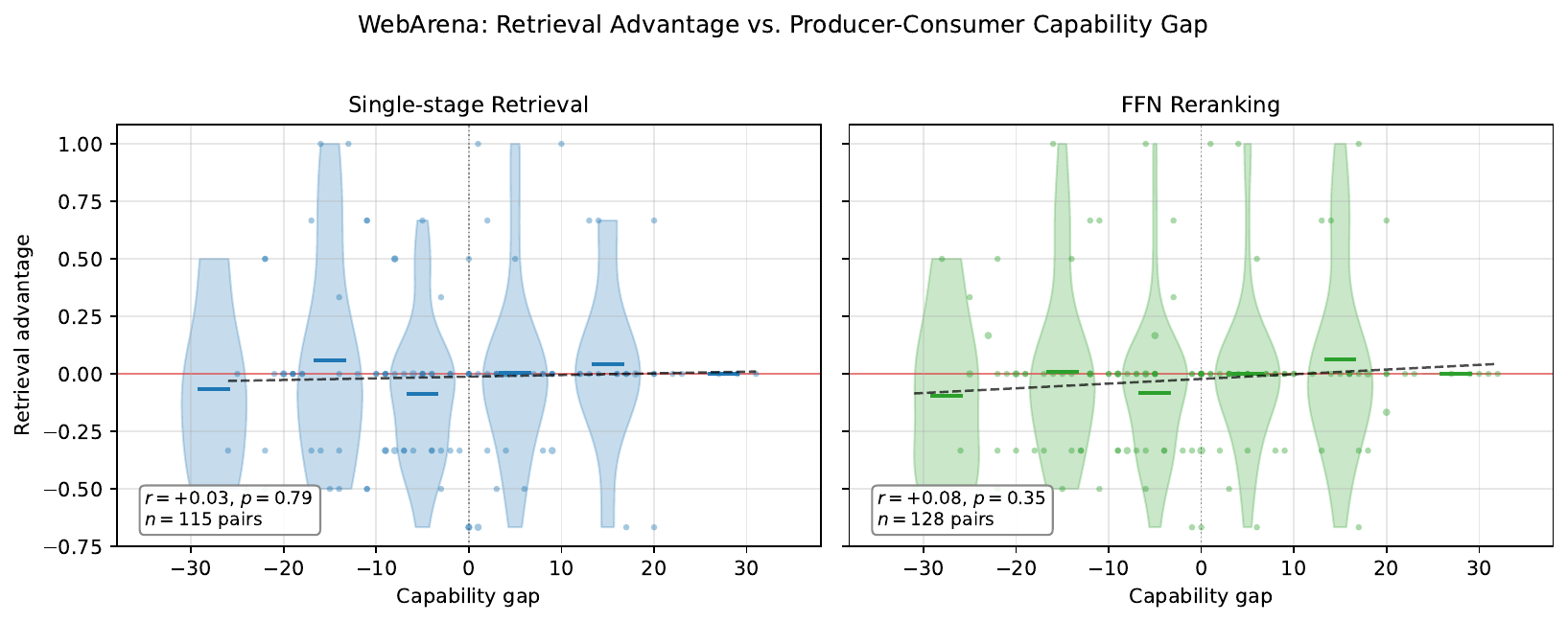}
  \caption{Retrieval Advantage vs.\ Producer-Consumer Capability Gap for ALFWorld (top) and WebArena (bottom).}
  \label{fig:rq3-app}
  % \vspace{-10pt}
\end{figure}

\subsection{Full results}
Figure~\ref{fig:rq3-app} shows $\mu_\alpha$ plotted against $\kappa_\alpha$ for ALFWorld and WebArena, with the best-performing reranker for each environment shown alongside the corresponding single-stage baseline.
Across both benchmarks and both retrieval settings, the Pearson correlation between $\kappa_\alpha$ and $\mu_\alpha$ is small and not statistically significant: $r = +0.04$
($p = 0.49$) for ALFWorld single-stage retrieval, $r = +0.09$
($p = 0.09$) for ALFWorld with SVMRank reranking, $r = +0.03$
($p = 0.79$) for WebArena single-stage retrieval, and $r = +0.08$
($p = 0.35$) for WebArena with FFN reranking.

Reranking consistently lifts the retrieval advantage distribution. On ALFWorld, the mean $\mu_\alpha$ rises from $+0.05$ under single-stage retrieval to $+0.1$ under SVMRank, and the fraction of pairs with $\mu_\alpha > 0$ rises from $51\%$ to $61\%$. 
On WebArena, the conditional means follow the same pattern: pairs where the producer is stronger than the consumer show a mean $\mu_\alpha$ of $+0.02$ under both single-stage and FFN reranking, while the overall distribution shifts upward under reranking. The correlation between $\kappa_\alpha$ and $\mu_\alpha$ also roughly doubles under reranking in both environments, from $r \approx 0.03$ to $r \approx 0.08$. 
While neither correlation reaches statistical significance, their consistency across two independent benchmarks and two retrieval settings indicates a real but small structural effect: reranking incorporates producer-capability information as one signal among many, consistent with the feature importance analysis in Appendix~\ref{sec:app:feature-importance}.

\section{Section \S\ref{subsec:rq4} Supplement}\label{sec:app:rq4-extra}

For ALFWorld, we adopt the six task types defined in the original benchmark; for WebArena, the 241 task intents. Because WebArena's task space is fine-grained, some test tasks have no same-type candidates in the index; we exclude such tasks from the WebArena experiments, leaving 47 tasks for this analysis. All results use the best-performing reranker per environment: SVMRank for ALFWorld and FFN for WebArena.
RPP in this section is the mean pairwise RPP among the three retrieval scopes (full, same task type, cross task type); the no-retrieval baseline is not part of this comparison set. Negative RPP values therefore indicate that a scope is dispreferred relative to the other two scopes on average.

\section{Section \S\ref{subsec:rq5} Supplement}\label{sec:app:rq5-full-results}
\begin{figure}[h]
  \centering
  \includegraphics[width=\columnwidth]{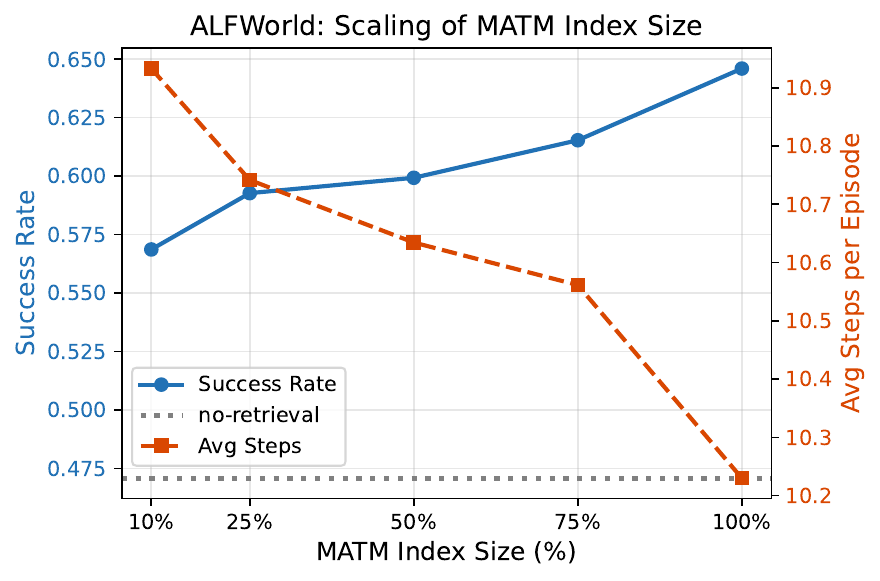}\\
  \includegraphics[width=\columnwidth]{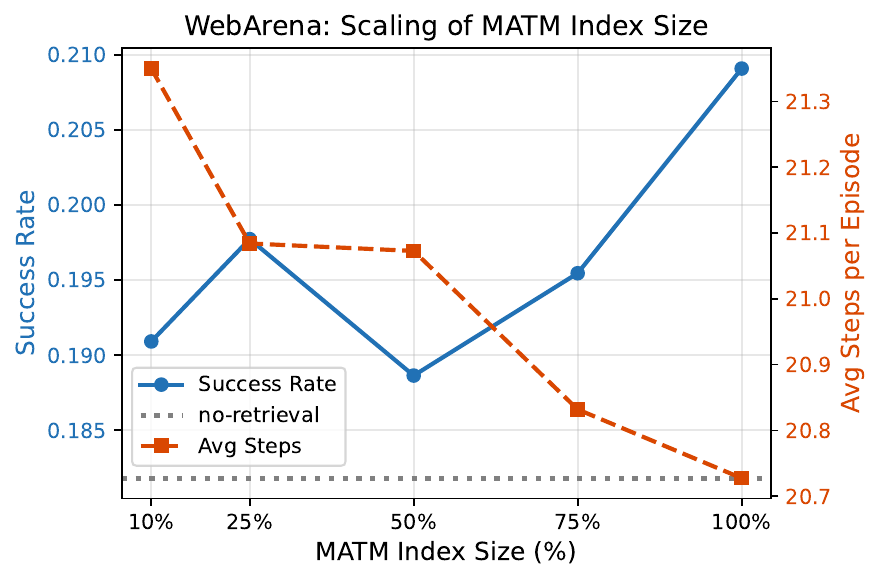}
  \caption{MATM memory scaling curves for ALFWorld (top) and WebArena (bottom). Success rate (left axis) and average steps per episode (right axis) as a function of index size. The dotted line marks SR of the no-retrieval baseline. Results are averaged over five runs with different random seeds.}
  \label{fig:app:rq5-scaling-curve}
\end{figure}

Figure~\ref{fig:app:rq5-scaling-curve} shows the full scaling curves for both environments. As discussed in Section~\ref{subsec:rq5}, ALFWorld exhibits monotonic improvement in both success rate and step efficiency with index size, confirming that larger memory consistently benefits the agent.

On WebArena, step efficiency also decreases monotonically, consistent with ALFWorld. Success rate, however, exhibits a non-monotonic pattern: it dips at the 50\% index before recovering sharply to 20.9\% at full scale, the strongest result across all index sizes and the clearest margin above the no-retrieval baseline of 18.2\%. 
We hypothesize that at intermediate scales, the index is large enough to surface plausible but ultimately unhelpful trajectories, yet not diverse enough to reliably include high-quality matches. 
At full scale, sufficient coverage overcomes this noise, restoring and exceeding the gains observed at smaller index sizes.

%<<<<<<<<New Edits
\section{Near-Duplicate Oracle Ceiling}\label{sec:app:oracle-ceiling}

\begin{table}[]
\centering
\small
\begin{tabular}{l c c c}
\hline
\textbf{Env} & \textbf{No Retr.} & \textbf{Oracle} & \textbf{Full} \\
 & (SR) & (SR) & (SR) \\
\hline
ALFWorld & 20.07 & \textbf{38.69} & 34.31 \\
WebArena & 16.58 & \textbf{40.38} & 19.97 \\
\hline
\end{tabular}
\caption{Near-duplicate oracle ceiling. Success rate (SR) for a single consumer agent (GPT-OSS 20B, single seed) when oracle trajectories (same task, same environment variation) are added to the corpus, compared against no retrieval and full retrieval. Oracle exposure approximates an upper bound.}
\label{tab:oracle-ceiling}
\end{table}

The retrieval-scope analysis in \S\ref{subsec:rq4} deliberately excludes \textit{oracle} trajectories---those solving the exact same task in the same environment variation as the test query---from the MATM corpus, since a producer having already solved the identical task instance is a rare, lucky event in an open agent population. To approximate an upper bound on retrieval benefit, we run an additional setting that reintroduces these oracle chunks into the corpus and compares it against no retrieval and full retrieval. Because of the additional cost, we use a single consumer agent (GPT-OSS 20B) with access to producer memory initialized from publicly available trajectories, and report a single random seed rather than the five-seed average used elsewhere.

Table~\ref{tab:oracle-ceiling} reports the results. Exposing oracle trajectories sharply increases success rate in both environments (ALFWorld $20.07 \to 38.69$; WebArena $16.58 \to 40.38$), well above full retrieval. This confirms that same-task/same-environment matches act as a performance ceiling: when a producer has solved the identical task instance, the consumer benefits substantially, which is a best case for MATM. Because such exact matches are rare in realistic open populations, we exclude oracle trajectories from the main corpus so that the reported gains reflect genuine cross-instance transfer rather than near-duplicate leakage.

\section{Token and Monetary Cost}\label{sec:app:token-cost}

\begin{table}[t]
\centering
\small
\resizebox{\columnwidth}{!}{
\begin{tabular}{l l c c c c}
\hline
\textbf{Model} & \textbf{Config} & \textbf{SR(\%)} $\uparrow$ & \textbf{Tokens/succ.} $\downarrow$ & \textbf{\$/succ.} $\downarrow$ & \textbf{Input tokens/ep.} \\
\hline
\multicolumn{6}{l}{\textbf{ALFWorld} (274 tasks)} \\
\hline
\multirow{2}{*}{GPT-OSS 20B}
& no retrieval & 25.18 & 118,181 & 0.00237 & 28,906 \\
& MATM & 40.15 & 78,629 (\textbf{$-$33.5\%}) & 0.00157 (\textbf{$-$33.9\%}) & 30,718 (+6.3\%) \\
\cdashline{1-6}
\multirow{2}{*}{Qwen3-32B}
& no retrieval & 37.96 & 73,188 & 0.00633 & 26,882 \\
& MATM & 54.74 & 52,949 (\textbf{$-$27.7\%}) & 0.00458 (\textbf{$-$27.6\%}) & 28,042 (+4.3\%) \\
\hline\hline
\multicolumn{6}{l}{\textbf{WebArena} (88 tasks)} \\
\hline
\multirow{2}{*}{GPT-OSS 20B}
& no retrieval & 13.64 & 1,539,334 & 0.02898 & 207,492 \\
& MATM & 22.73 & 1,012,911 (\textbf{$-$34.2\%}) & 0.01901 (\textbf{$-$34.4\%}) & 227,753 (+9.8\%) \\
\cdashline{1-6}
\multirow{2}{*}{Qwen3-32B}
& no retrieval & 17.05 & 1,126,126 & 0.09223 & 190,133 \\
& MATM & 18.18 & 1,005,794 (\textbf{$-$10.7\%}) & 0.08240 (\textbf{$-$10.7\%}) & 181,111 ($-$4.7\%) \\
\hline
\end{tabular}
}
\caption{Token usage and monetary cost of single consumer agents without retrieval and with single-stage MATM retrieval. `Tokens/succ.' and `\$/succ.' divide total tokens (input and output) and total cost by the number of successful episodes. Percentages in parentheses are changes relative to no retrieval in the same setting.}
\label{tab:token-cost}
\end{table}

Fewer interaction steps (\S\ref{subsec:rq1}) may not imply lower inference cost, because every retrieval call adds a trajectory chunk to the consumer agent's prompt. We therefore measure token usage and monetary cost directly for two consumer agents, GPT-OSS 20B and Qwen3-32B, each run on its own on the full test sets (274 ALFWorld and 88 WebArena tasks) without retrieval and with single-stage MATM retrieval.

For every action-generation call, we count input tokens as the system and user prompt rendered through the model's chat template, and output tokens as the generated reasoning and action, using each model's own tokenizer. Costs use OpenRouter prices per million tokens (\$0.018 input / \$0.09 output for GPT-OSS 20B; \$0.08 / \$0.28 for Qwen3-32B). Since a configuration that spends more per attempted episode can still be cheaper per task it actually solves, we normalize by successful episodes: tokens per success divides the total number of input and output tokens by the number of successful episodes, and dollars per success divides the total cost in the same way.

Table~\ref{tab:token-cost} reports the results. Input tokens per episode increase slightly with retrieval in three of the four settings, since the retrieved trajectory lengthens the prompt and fewer steps only partly offset it. Per solved task, however, MATM reduces token usage in every setting, by 33.5\% and 27.7\% on ALFWorld and by 34.2\% and 10.7\% on WebArena for GPT-OSS 20B and Qwen3-32B respectively, with nearly identical reductions in dollar cost. Input tokens account for 88--97\% of the total cost in all settings, so retrieval's effect on cost is driven by prompt length rather than generation length. One possible explanation for the input token decrease for Qwen3-32B on WebArena is that its prompts are dominated by the page's accessibility tree, so a shift toward smaller pages under retrieval can outweigh the added trajectory chunk.

\section{Reproducibility}\label{sec:app:reproducibility}

We release all resources needed to reproduce our results.
\begin{itemize}
    \item \textbf{Code and README.} Full training, indexing, and evaluation code with reproduction instructions: \url{https://github.com/kimdanny/matm}.
    \item \textbf{Trajectory dataset.} The complete MATM trajectory corpus: \url{https://huggingface.co/datasets/toeunkim/matm-trajectories}.
    \item \textbf{Trained rerankers.} The trained LTRT reranker models are included in the code repository.
\end{itemize}
%<<<<<<<<New Edits (end)

\section{Dataset License}
\begin{itemize}
    \item ALFWorld: MIT License
    \item WebArena: Apache License 2.0
\end{itemize}

\section{Computational Budget}
For retrieval from the MATM index, we use one NVIDIA L40S GPU.
For LLM inference in experiments, the OpenRouter API was used. The total cost was approximately 2,000 USD.

\section{Use of AI Assistants}
AI assistants were used for paraphrasing during paper writing and for simple implementation tasks during coding. All outputs were thoroughly reviewed by the authors.

\onecolumn
\section{Language Model Prompts}\label{sec:app:prompts}
%%%%%%%%%%%%%%% 
\subsection{Retrieval Planner Prompt}\label{subsec:app:retrieval-planner}

\begin{systemprompt}
You are a retrieval planner for an agent. Decide if a new retrieval of a successful trajectory is needed now. Consider the goal, recent observations, and the current retrieved trajectory context. Focus on identifying if the current situation is similar to the retrieved trajectory. If the current situation and recent history is similar to the retrieved trajectory, return [NoRetrieval]. If the current situation and recent history is not similar to the retrieved trajectory, return [Retrieval].
\end{systemprompt}
\begin{userprompt}
\{ \\
\quad "goal": "<goal\_text>", \\
\quad "current\_observation": "<observation>", \\
\quad "recent\_history": "<recent\_history\_str>", \\
\quad "current\_retrieved\_context": "<trajectory\_context>" \\
\} \\[0.5em]
Return only one of these exact tags: [Retrieval] or [NoRetrieval]. Do not include any other text.
\end{userprompt}
%%%%%%%%%%%%%%% 

%%%%%%%%%%%%%%% ALFWorld
\subsection{ALFWorld Baseline (no-retrieval) Prompt}
Note that the `ONE-SHOT EXAMPLE' used in this section is an illustrative example; the prompts are adjusted based on the task.

\begin{systemprompt}
    Interact with a household to solve a task. Imagine you are an intelligent agent in a household environment and your target is to perform actions to complete the task goal. At the beginning of your interactions, you will be given the detailed description of the current environment and your goal to accomplish. For each of your turn, you will be given the observation of the last turn. Think step by step about what you observe and what action to take next. The available actions are: 1. go to recep 2. take obj from recep 3. put obj in/on recep 4. open recep 5. close recep 6. toggle obj recep 7. clean obj with recep 8. heat obj with recep 9. cool obj with recep where obj and recep correspond to objects and receptacles. After your each turn, the environment will give you immediate feedback based on which you plan your next few steps. if the envrionment output "Nothing happened", that means the previous action is invalid and you should try more options. Reminder: 1. The action must be chosen from the given available actions. Any actions except provided available actions will be regarded as illegal. 2. Think when necessary, try to act directly more in the process. \\[1em]
    RESPONSE FORMAT: \\[0.5em]
    For each turn, you must provide your response in JSON format with two fields: \\[0.5em]
    - reasoning: MUST start with `Let's think step by step.' followed by detailed reasoning about the situation and why this action is the best choice \\[0.5em]
    - action: your chosen action (must be exactly from admissible\_actions list) \\[0.5em]
    CRITICAL RULES: \\[0.5em]
    1. Your action MUST be character-for-character identical to one item in admissible\_actions \\[0.5em]
    2. Do NOT modify, abbreviate, or paraphrase actions \\[0.5em]
    3. Do NOT use actions from retrieved trajectories unless they appear in current admissible\_actions \\[0.5em]
    4. If confused, pick a safe exploratory action like "look" or navigation \\[1em]
    
    ONE-SHOT EXAMPLE: \\[0.5em]
    Task: clean some apple and put it in sidetable. \\[1em]
    Turn 1: \\[0.5em]
    Observation: The fridge 1 is closed. \\[0.5em]
    Response: \{"reasoning": "Let's think step by step. I should open fridge 1 to see what's inside.", "action": "open fridge 1"\} \\[1em]
    Turn 2: \\[0.5em]
    Observation: You open the fridge 1. The fridge 1 is open. In it, you see a lettuce 2, a mug 2, a potato 2, and a tomato 1. \\[0.5em]
    Response: \{"reasoning": "Let's think step by step. I need to navigate to diningtable 1 to continue searching or complete the task.", "action": "go to diningtable 1"\} \\[1em]
    Turn 3: \\[0.5em]
    Observation: On the diningtable 1, you see nothing. \\[0.5em]
    Response: \{"reasoning": "Let's think step by step. I need to navigate to diningtable 2 to continue searching or complete the task.", "action": "go to diningtable 2"\} \\

    [TURNS SKIPPED FOR READABILITY]\\

    Turn 11: \\[0.5em]
    Observation: On the sinkbasin 1, you see a cup 2, a egg 1, a mug 1, and a potato 3. \\[0.5em]
    Response: \{"reasoning": "Let's think step by step. I need to clean this object as required by the task.", "action": "clean apple 3 with sinkbasin 1"\} \\[1em]
    Turn 12: \\[0.5em]
    Observation: You clean the apple 3 using the sinkbasin 1. \\[0.5em]
    Response: \{"reasoning": "Let's think step by step. Now I clean an apple (3). Next, I need to put it in/on sidetable 1.", "action": "go to sidetable 1"\} \\[1em]
    Turn 13: \\[0.5em]
    Observation: On the sidetable 1, you see a cup 1, a lettuce 1, a peppershaker 3, a potato 1, and a saltshaker 1. \\[0.5em]
    Response: \{"reasoning": "Let's think step by step. Now I have the object, I need to place it at the target location to complete the task.", "action": "put apple 3 in/on sidetable 1"\}
\end{systemprompt}
\begin{userprompt}
    GOAL: <goal\_text> \\[1em]
    CURRENT STEP: <current\_step> / <max\_steps> \\[1em]
    --- RECENT HISTORY (Previous Steps - For Reference Only) --- \\[0.5em]
    <recent\_history\_str> \\[0.5em]
    --- End of Recent History --- \\[1em]
    >{}>{}> CURRENT OBSERVATION (Focus on This - Current State): \\[0.5em]
    <observation> \\[0.5em]
    <{}<{}< End of Current Observation \\[1em]
    ADMISSIBLE ACTIONS (<N> total): \\[0.5em]
    1. <action\_1> \\[0.5em]
    2. <action\_2> \\[0.5em]
    ... \\[1em]
    RESPONSE FORMAT: \\[0.5em]
    You MUST respond with valid JSON in this exact format: \\[0.5em]
    \{"reasoning": "Let's think step by step. [your detailed reasoning]", "action": "exact action from admissible\_actions"\} \\[1em]
    Where: \\[0.5em]
    - reasoning: MUST start with `Let's think step by step.' Then explain your thought process, what you observe, and why this action is best \\[0.5em]
    - action: Must be EXACTLY one string from the admissible\_actions list above (character-for-character match) \\[1em]
    IMPORTANT: \\[0.5em]
    1. Your reasoning MUST begin with `Let's think step by step.' \\[0.5em]
    2. Do not include any text before or after the JSON object.
\end{userprompt}

\subsection{ALFWorld Trajectory-Augmented Prompt}
\begin{systemprompt}
        Same as \textit{ALFWorld Baseline (no-retrieval) System Prompt}
\end{systemprompt}
\begin{userprompt}
    GOAL: <goal\_text> \\[1em]
    CURRENT STEP: <current\_step> / <max\_steps> \\[1em]
    --- RECENT HISTORY (Previous Steps - For Reference Only) --- \\[0.5em]
    <recent\_history\_str> \\[0.5em]
    --- End of Recent History --- \\[1em]
    --- RETRIEVED TRAJECTORY GUIDANCE (Reference Examples) --- \\[0.5em]
    RETRIEVED TRAJECTORY: \\[0.5em]
    Task: <task\_description> \\[1em]
    Retrieved successful trajectory sequence: \\[0.5em]
    Step 1: <action\_1> \\[0.5em]
      Observation: <observation\_1> \\[0.5em]
    Step 2: <action\_2> \\[0.5em]
      Observation: <observation\_2> \\[0.5em]
    ... \\[1em]
    Use this trajectory as a reference for your planning. Consider: \\[0.5em]
    1. The sequence of actions that led to success \\[0.5em]
    2. The observations and their progression \\[0.5em]
    3. How to adapt this strategy to the current situation and your goal \\[0.5em]
    4. What steps might be different or similar in your current context \\[0.5em]
    --- End of Trajectory Guidance --- \\[1em]
    >{}>{}> CURRENT OBSERVATION (Focus on This - Current State): \\[0.5em]
    <observation> \\[0.5em]
    <{}<{}< End of Current Observation \\[1em]
    ADMISSIBLE ACTIONS (<N> total): \\[0.5em]
    1. <action\_1> \\[0.5em]
    2. <action\_2> \\[0.5em]
    ... \\[1em]
    RESPONSE FORMAT: \\[0.5em]
    You MUST respond with valid JSON in this exact format: \\[0.5em]
    \{"reasoning": "Let's think step by step. [your detailed reasoning]", "action": "exact action from admissible\_actions"\} \\[1em]
    Where: \\[0.5em]
    - reasoning: MUST start with `Let's think step by step.' Then explain your thought process, what you observe, and why this action is best \\[0.5em]
    - action: Must be EXACTLY one string from the admissible\_actions list above (character-for-character match) \\[1em]
    IMPORTANT: \\[0.5em]
    1. Your reasoning MUST begin with `Let's think step by step.' \\[0.5em]
    2. Do not include any text before or after the JSON object.
\end{userprompt}

%%%%%%%%%%%%%%% WebArena
\subsection{WebArena Baseline (no-retrieval) Prompt}
\begin{systemprompt}
    You are an autonomous intelligent agent tasked with navigating a web browser. You will be given web-based tasks. These tasks will be accomplished through the use of specific actions you can issue. \\[1em]
    Here's the information you'll have: \\[0.5em]
    The user's objective: This is the task you're trying to complete. \\[0.5em]
    The current web page's accessibility tree: This is a simplified representation of the webpage, providing key information. \\[0.5em]
    The current web page's URL: This is the page you're currently navigating. \\[0.5em]
    The open tabs: These are the tabs you have open. \\[0.5em]
    The previous action: This is the action you just performed. It may be helpful to track your progress. \\[1em]
    The actions you can perform fall into several categories: \\[1em]
    Page Operation Actions: \\[0.5em]
    `click [id]`: This action clicks on an element with a specific id on the webpage. \\[0.5em]
    `type [id] [content] [press\_enter\_after=0|1]`: Use this to type the content into the field with id. By default, the "Enter" key is pressed after typing unless press\_enter\_after is set to 0. \\[0.5em]
    `hover [id]`: Hover over an element with id. \\[0.5em]
    `press [key\_comb]`: Simulates the pressing of a key combination on the keyboard (e.g., Ctrl+v). \\[0.5em]
    `scroll [direction=down|up]`: Scroll the page up or down. \\[1em]
    Tab Management Actions: \\[0.5em]
    `new\_tab`: Open a new, empty browser tab. \\[0.5em]
    `tab\_focus [tab\_index]`: Switch the browser's focus to a specific tab using its index. \\[0.5em]
    `close\_tab`: Close the currently active tab. \\[1em]
    URL Navigation Actions: \\[0.5em]
    `goto [url]`: Navigate to a specific URL. \\[0.5em]
    `go\_back`: Navigate to the previously viewed page. \\[0.5em]
    `go\_forward`: Navigate to the next page (if a previous 'go\_back' action was performed). \\[1em]
    Completion Action: \\[0.5em]
    `stop [answer]`: Issue this action when you believe the task is complete. If the objective is to find a text-based answer, provide the answer in the bracket. \\[1em]
    Available Websites: \\[0.5em]
    You have access to the following websites: \\[0.5em]
    - OneStopShop (E-commerce): <SHOPPING\_URL> \\[0.5em]
    - Shopping Admin (CMS): <SHOPPING\_ADMIN\_URL> \\[0.5em]
    - Reddit (Forum): <REDDIT\_URL> \\[0.5em]
    - GitLab: <GITLAB\_URL> \\[0.5em]
    - Wikipedia: <WIKIPEDIA\_URL> \\[0.5em]
    - OpenStreetMap: <MAP\_URL> \\[1em]
    RESPONSE FORMAT: \\[0.5em]
    For each turn, you must provide your response in JSON format with two fields: \\[0.5em]
    - reasoning: Start with 'Let's think step by step' and then provide detailed step-by-step reasoning about the current situation, what you observe, and why the chosen action is the best next step \\[0.5em]
    - action: your chosen action (must be exactly from the available actions) \\[1em]
    IMPORTANT: \\[0.5em]
    1. You should only issue an action that is valid given the current observation \\[0.5em]
    2. You should only issue one action at a time \\[0.5em]
    3. For each observation, think step by step about what you see and what action to take next \\[0.5em]
    4. Your action MUST be one of the valid action formats listed above \\[1em]
    \textbf{Few-shot examples and closing sentence.} \\[0.5em]
    Example 1: \\[0.5em]
    OBSERVATION: \\[0.5em]
    [1744] link 'HP CB782A\#ABA 640 Inkjet Fax Machine (Renewed)' \\[0.5em]
    [1749] StaticText '\$279.49' \\[0.5em]
    [1757] button 'Add to Cart' \\[0.5em]
    [1760] button 'Add to Wish List' \\[0.5em]
    [1761] button 'Add to Compare' \\[0.5em]
    URL: http://onestopmarket.com/office-products/office-electronics.html \\[0.5em]
    OBJECTIVE: What is the price of HP CB782A\#ABA 640 Inkjet Fax Machine (Renewed)? \\[0.5em]
    PREVIOUS ACTION: None \\[0.5em]
    Response: \{"reasoning": "Let's think step by step. I can see the product HP CB782A\#ABA 640 Inkjet Fax Machine (Renewed) is displayed on the page. The price is clearly shown as \$279.49 in element [1749]. Since the objective is to find the price, and I have found it, I should stop and provide the answer.", "action": "stop [\$279.49]"\} \\[1em]
    Example 2: \\[0.5em]
    OBSERVATION: \\[0.5em]
    [164] textbox 'Search' focused: True required: False \\[0.5em]
    [171] button 'Search' \\[0.5em]
    URL: http://onestopmarket.com \\[0.5em]
    OBJECTIVE: Search for USB-C to USB-B cable on the site. \\[0.5em]
    PREVIOUS ACTION: None \\[0.5em]
    Response: \{"reasoning": "Let's think step by step. I need to search for USB-C to USB-B cable on this site. I can see a search textbox with id [164] that is already focused. I should type my search query into this textbox and press enter to execute the search. I'll use press\_enter\_after=1 to submit the search after typing.", "action": "type [164] [USB-C to USB-B cable] [1]"\} \\[1em]
    Example 3: \\[0.5em]
    OBSERVATION: \\[0.5em]
    [1124] heading 'Certified Refurbished Kindle Paperwhite' \\[0.5em]
    [1136] StaticText 'by Amazon' \\[0.5em]
    [1163] StaticText 'Price: \$79.99' \\[0.5em]
    [1189] button 'Add to Cart' \\[0.5em]
    URL: http://onestopmarket.com/dp/B00ZV 9RDKK \\[0.5em]
    OBJECTIVE: Show me the details of the Kindle Paperwhite. \\[0.5em]
    PREVIOUS ACTION: click [582] \\[0.5em]
    Response: \{"reasoning": "Let's think step by step. I can see the Kindle Paperwhite product page with the heading, price (\$79.99), and an Add to Cart button. The objective is to show details of the product. Currently, I can only see basic information. To view more detailed information like specifications, features, or description, I should scroll down the page to see if there are additional details below.", "action": "scroll [down]"\} \\[1em]
    Now, given the current observation, think step-by-step and provide your response.
\end{systemprompt}
\begin{userprompt}
    GOAL: <goal\_text> \\[1em]
    CURRENT STEP: <current\_step> / <max\_steps> \\[1em]
    --- RECENT HISTORY (Previous Steps - For Reference Only) --- \\[0.5em]
    <recent\_history\_str> \\[0.5em]
    --- End of Recent History --- \\[1em]
    >{}>{}> CURRENT OBSERVATION (Focus on This - Current State): \\[0.5em]
    <observation> \\[0.5em]
    <{}<{}< End of Current Observation \\[1em]
    CURRENT URL: <url> \\[1em]
    RESPONSE FORMAT: \\[0.5em]
    You MUST respond with valid JSON in this exact format: \\[0.5em]
    \{"reasoning": "Let's think step by step. [your detailed reasoning]", "action": "action to be taken"\} \\[1em]
    Where: \\[0.5em]
    - reasoning: MUST start with `Let's think step by step.' Then explain your thought process, what you observe, and why this action is best \\[0.5em]
    - action: action to be taken \\[1em]
    IMPORTANT: \\[0.5em]
    1. Your reasoning MUST begin with `Let's think step by step.' \\[0.5em]
    2. Do not include any text before or after the JSON object.
\end{userprompt}

\subsection{WebArena Trajectory-Augmented Prompt}
\begin{systemprompt}
        Same as \textit{WebArena Baseline (no-retrieval) System Prompt} 
\end{systemprompt}
\begin{userprompt}
    GOAL: <goal\_text> \\[1em]
    CURRENT STEP: <current\_step> / <max\_steps> \\[1em]
    --- RECENT HISTORY (Previous Steps - For Reference Only) --- \\[0.5em]
    <recent\_history\_str> \\[0.5em]
    --- End of Recent History --- \\[1em]
    --- RETRIEVED TRAJECTORY GUIDANCE (Reference Examples) --- \\[0.5em]
    RETRIEVED TRAJECTORY: \\[0.5em]
    Task: <task\_description> \\[1em]
    Retrieved successful trajectory: \\[0.5em]
    Step 1: \\[0.5em]
      Action: <action\_1> \\[0.5em]
      Observation: <observation\_1> \\[0.5em]
    Step 2: \\[0.5em]
      Action: <action\_2> \\[0.5em]
      Observation: <observation\_2> \\[0.5em]
    ... \\[1em]
    Use this trajectory as a reference for your planning. Consider: \\[0.5em]
    1. The sequence of actions taken \\[0.5em]
    2. How the agent navigated through the website \\[0.5em]
    3. What elements were clicked and in what order \\[0.5em]
    4. When the task was completed \\[0.5em]
    --- End of Trajectory Guidance --- \\[1em]
    >{}>{}> CURRENT OBSERVATION (Focus on This - Current State): \\[0.5em]
    <observation> \\[0.5em]
    <{}<{}< End of Current Observation \\[1em]
    CURRENT URL: <url> \\[1em]
    RESPONSE FORMAT: \\[0.5em]
    You MUST respond with valid JSON in this exact format: \\[0.5em]
    \{"reasoning": "Let's think step by step. [your detailed reasoning]", "action": "action to be taken"\} \\[1em]
    Where: \\[0.5em]
    - reasoning: MUST start with `Let's think step by step.' Then explain your thought process, what you observe, and why this action is best \\[0.5em]
    - action: action to be taken \\[1em]
    IMPORTANT: \\[0.5em]
    1. Your reasoning MUST begin with `Let's think step by step.' \\[0.5em]
    2. Do not include any text before or after the JSON object.
\end{userprompt}

\end{document}